\documentclass{egpubl}

\JournalPaper         
\usepackage[T1]{fontenc}
\usepackage{dfadobe}

\biberVersion
\BibtexOrBiblatex
\usepackage[backend=biber,bibstyle=EG,citestyle=alphabetic,backref=true]{biblatex} 
\electronicVersion
\PrintedOrElectronic

\ifpdf \usepackage[pdftex]{graphicx} \pdfcompresslevel=9
\else \usepackage[dvips]{graphicx} \fi

\usepackage{egweblnk} 

\usepackage{graphicx}
\usepackage{comment}
\usepackage{amsmath,amssymb} 
\usepackage{xcolor}
\usepackage{url}
\usepackage{microtype}
\usepackage{rotating}
\usepackage{booktabs}
\usepackage{arydshln}
\usepackage{subcaption}
\usepackage{soul}
\usepackage{epstopdf}

\makeatletter
\newcommand{\printfnsymbol}[1]{%
  \textsuperscript{\@fnsymbol{#1}}%
}
\makeatother

\DeclareMathOperator{\E}{\mathbb{E}}

\newcommand*\samethanks[1][\value{footnote}]{\footnotemark[#1]}

\title[EG \LaTeX\ Author Guidelines]%
      {\LaTeX\ Author Guidelines for EUROGRAPHICS Proceedings Manuscripts}

\author[S. Benaim, R. Mokady, A. Bermano , L. Wolf]
{\parbox{\textwidth}{\centering S\
. Benaim\thanks{Equal contribution.}$^{1}$
, R. Mokady\samethanks$^{1}$, A. Bermano$^{1}$, L. Wolf$^{1,2}$\thanks{E-mails: [sagieb@tauex.tau.ac.il; ronmokady@mail.tau.ac.il; amberman@tauex.tau.ac.il; wolf@cs.tau.ac.il] } 
        }
        \\
{\parbox{\textwidth}{\centering $^1$ The School of Computer Science, Tel Aviv University\\
         $^2$ Facebook AI Research
       }
}
}

\begin{document}

\teaser{
 \includegraphics[width=0.7\linewidth]{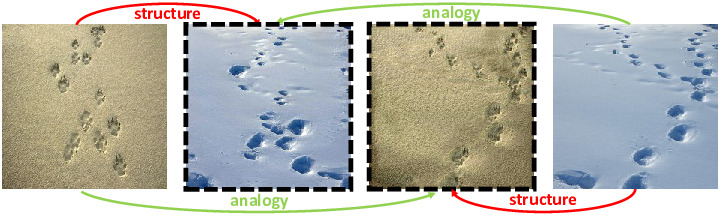}
 \centering
  \caption{Our method takes two images as input (left and right), and generates images that consist of features from one image, spatially structured analogically to the other.}
\label{fig:teaser}
}

\ccsdesc[300]{Computing methodologies~Image representations}

\title{Structural Analogy from a Single Image Pair}

\maketitle
\begin{abstract} 
The task of unsupervised image-to-image translation has seen substantial advancements in recent years through the use of deep neural networks. Typically, the proposed solutions learn the characterizing distribution of two large, unpaired collections of images, and are able to alter the appearance of a given image, while keeping its geometry intact. In this paper, we explore the capabilities of neural networks to understand image \textit{structure} given only a single pair of images, $A$ and $B$. We seek to generate images that are \textit{structurally aligned}: that is, to generate an image that keeps the appearance and style of $B$, but has a structural arrangement that corresponds to $A$. The key idea is to map between image patches at different scales. This enables controlling the granularity at which analogies are produced, which determines the conceptual distinction between style and content. In addition to \textit{structural alignment}, our method can be used to generate high quality imagery in other conditional generation tasks utilizing images $A$ and $B$ only: guided image synthesis, style and texture transfer, text translation as well as video translation. Our code and additional results are available in \url{https://github.com/rmokady/structural analogy/}.

\end{abstract}

\section{Introduction}

The task of image-to-image translation has seen tremendous growth in recent years~\cite{pix2pix,pix2pixHD,spade,bicyclegan,CycleGAN2017,discogan,dualgan,cogan,unit,stargan,conneau2017word,zhang2017adversarial,zhang2017earth,lample2018unsupervised, DRIT_plus}. In the typical setting, an image is generated such that it depicts the same scene as in an input source image, but displays the visual properties of a target one. At first, supervision was in the form of explicit matching image pairs in both domains, illustrating the exact same content~\cite{pix2pix}. These sorts of examples are hard to obtain, and hence researchers searched for ways to relax this requirement, yielding a series of \textit{unpaired} translation methods. In this satisfying setting, a large collection of images is still required for each domain, but they do not need to be matched~\cite{CycleGAN2017,discogan,dualgan,cogan,unit}. 

Recent works, however, have demonstrated that a lot of information can already be extracted from just a single image, due to the information residing within the internal statistics of patches comprising the image~\cite{DIP,DoubleDIP,Zhou2018,ingan,singan}. While very insightful and inspiring, these works are restricted to learning the said distribution of a single image, and are not appropriate to the image-to-image translation task (see Section~\ref{sec:related} for a more comprehensive discussion). 

The literature can also be organized by the distinction between content and style, although these concepts are not necessarily disjointed. When translating a bundle of balloons to a flock of birds, should the shape of each balloon be preserved? should the arrangement within the bundle? Should color stripes within a balloon be shaped like feathers?  
After successfully translating painting styles, textures~\cite{pix2pix, CycleGAN2017}, and environment illumination~\cite{unit, munit}, recent work has experimented with changing shapes in the image, along with texture~\cite{domainintersection}. Other work considers the case of geometry altering transformations~\cite{deformation, transgaga}.

In this paper, we present the first work to combine both of the discussed concepts, i.e. performs image-to-image translation from a single pair, and aims at learning structure along with appearance. Specifically, our work looks for \textit{structural analogies}, generating an image that is: (1) similarly \textit{aligned}: i.e. it follows the spatial distribution, or \textit{structure}, of features from the target image, and (2) \textit{analogous}: i.e. it depicts features that correlate to the target structure, but are from the source image. This is in contrast, for example, to Deep Image Analogy~\cite{deepimageanalogy}, which looks for analogous image positions,  effectively transferring appearance between the images while preserving shapes.

We follow the already established notion of ``structure" in works such as Deep Image Analogy~\cite{deepimageanalogy} and SinGAN~\cite{singan}. This notion extends the traditional concept of ``style", by representing, for each scale, the spatial distribution of the objects and their inter-relations, disregarding finer details, such as texture. For example, the global structure is captured at coarse scales, similarly to SinGAN~\cite{singan}, e.g. sky at the top. 
The medium scales contribute to the spatial distribution of different objects, such as the hot-air balloons of various colors, including their relative positions to each other, and absolute location. Lastly, at the finest scale, texture resides.

In our case, performing the analogy on only small scale features induces traditional appearance transfer, but the more scales we add, the larger the features that are being replaced during the translation. 
In Fig.~\ref{fig:teaser} an example of our method is shown, where the appearance is translated (e.g. sand becomes snow), but in addition larger features, such as the footprints, are also translated as to their counterparts nicely and fully. This is done with no supervision directing the translation to keep the semantic structure intact, except for a similarity loss on image patches.

At each scale, our method first generates a sample of the source domain, conditioned by the previous scale, and a random input, in a similar fashion to the previously proposed SinGAN~\cite{singan}. Then, using a cycle consistency process~\cite{CycleGAN2017, discogan, dualgan}, the sample is translated to the target domain, where it is expected to match the target image. 
Our work extends style-transfer methods: Using the finest levels of our generative hierarchy, we can perform traditional style-transfer. Using more levels, our method identifies structure at different scales and finds more meaningful analogies.
The driving concept is that for a correctly structured (or \textit{well aligned}) sample, a shape preserving style transfer would produce an image that is similar to the target one at every corresponding scale.

Existing content-style disentanglement works, such as DIA~\cite{deepimageanalogy}, Jha et al.~\cite{jha2018disentangling}, Denton and Birodkar~\cite{denton2017unsupervised}, Bouchacourt et al.~\cite{bouchacourt},  Gabbay and Hoshen~\cite{gabbay} and style transfer works such as Gatys et al.~\cite{Gatys_2016_CVPR} and Huang and Belongie~\cite{adain}, are unable to change the geometry of objects when generating analogies. Our method can change the structure of the objects, resulting in both aligned and realistic transformations. 
DIA~\cite{deepimageanalogy}, for example, cannot faithfully map between birds and balloons, as their geometric structure is completely different (see Sec.~\ref{sec:visual_results} for details). 
Many content-style disentanglement works such as Jha et al.~\cite{jha2018disentangling}, Denton and Birodkar~\cite{denton2017unsupervised}, Bouchacourt et al.~\cite{bouchacourt},  Gabbay and Hoshen ~\cite{gabbay} rely on class level supervision which is not available in our case.

To summarize, our paper is the first to introduce the problem of structural analogies from a single image pair. We introduce a key idea, inspired by classical works such as that of Simakov et al.,~\cite{michals}: a bi-directional mapping of multi-scale patch distributions between two images preserves strong structural semantics. Extending this idea, we identify and employ “structural analogies”, i.e. we generate a new image such that for each scale, the spatial arrangement of objects is preserved. We then provide a novel method to realize these concepts. 

We demonstrate our method on the task of \textit{structural alignment}, as well the conditional generation tasks of guided image synthesis, style and texture transfer and texture synthesis. We achieve comparable or favorable results to the state-of-the-art, demonstrating that unsupervised image-to-image translation from only two images is indeed possible, thanks to the powerful tool of multi-scale structural analogy, and information residing in patch statistics of single images. 

\section{Related Work}
\label{sec:related}

In recent years, the emergence of Generative Adversarial Networks (GANs)~\cite{gan} has changed the landscape of many fields, and in particular, led to an intense advancement in generative image modeling. GANs can learn how to  generate images with a distribution that closely resembles the one of the training set. Nevertheless, unconditional GANs are still quite limited and require a huge amount of training data to achieve plausible visual results~\cite{biggan}.

Conditional GANs on the other hand, generally allows higher fidelity and yield higher visual quality. The most notable advancement has been in the development of image-to-image translation methods, which dramatically revolutionized many applications. Pix2Pix~\cite{pix2pix} have presented a generic method that learns to translate between two domains. Their method is based on a conditional GAN that learns from (supervised) paired data. They utilize both a conditional GAN loss to generate realistic looking images and $L1$ loss to match the ground truth image. Pix2PixHD~\cite{pix2pixHD} uses two separate low-res and high-res networks to generate images of high resolution. SPADE~\cite{spade} performs semantic image synthesis by introducing a new type of conditional normalization. BicycleGAN~\cite{bicyclegan}, as another example, injects random noise into the generator and attempts to recover the noise from the generated output. All of these works, however, require costly annotation of matching pairs of images, which is unfeasible for many applications. 

Learning to translate from (unsupervised) unpaired data is more challenging.  Taigman et al.~\cite{taigman} presented a method based on a conditional GANs to transfer between unpaired domains, using a perceptual loss that is based on a classifier. CycleGAN~\cite{CycleGAN2017}, DiscoGAN~\cite{discogan} and DualGAN~\cite{dualgan} presented generic networks with cycle-consistency that learn to translate between two given unpaired sets without additional supervision. These works have lead the way to many unpaired image-to-image translation networks~\cite{unit, munit, DRIT_plus, stgan}. Nevertheless, these techniques mainly transfer the appearance (or style) from one image to another, and struggle in transferring the shape or the structure, as required for creating an analogy that is truly faithful to the target domain. 

A number of recent works have challenged this problem using various means. For example, Benaim et al.~\cite{domainintersection} use a disentangled latent space to map domain-specific content. Wu et al.~\cite{transgaga} propose disentangling the space of images into a Cartesian product of appearance and geometry latent spaces to allow for a geometry-aware translation. Katzir et al.~\cite{katzir} propose applying the translation on deep feature maps of a VGG network. These approaches require a large collection of images, stratified in a way that exposes the structural information.

Recently, a number of works~\cite{Zhou2018,ingan,singan} presented 
GANs capable of learning and generating from  the internal patch distribution of a single image. However, since the patch statistics are not trained in the context of another image, such single sample models cannot effectively map between two very different distributions. While technically, one can force such models to perform such a mapping, our experiments show that this leads to poor outcomes.

Our work is also related to style transfer methods~\cite{Efros2000, Hertzmann2000, li2017universal, luan2017deep, sunkavalli2010multi, luan2018deep}. The most notable work based on neural network is of Gatys et al.~\cite{Gatys_2016_CVPR} and Huang and Belongie~\cite{adain} which attempt to transfer the style or a source image to a target one. When applied at a very fine scale, our work compares with such methods, using a completely different technique. A specific work in the context of text stylization by example~\cite{font1} maps a texture to a binary map of a letter or word. This is a much more limiting scenario than ours. Even though, as we demonstrate, our method is also able to produce pleasing images of stylized text.
Concurrently to our work, \cite{tuigan} attempts to learn transformation between two
pairs of images using an hierarchical model.

Deep image analogies~\cite{deepimageanalogy} is apparently the closest work to ours, since they use a single pair of images to transfer style. Unlike us, they assume that the two input images share similar semantics. By analyzing their VGG deep features, they mix content features and style features to synthesize novel images. Conceptually, their work is significantly different than ours, since they cannot change the structure of the image, moreover, their method is not a generative model, and no novel patches are generated. Lastly, we do not rely on a pre-trained classifier, which is a form of supervision.

In a similar manner to DIA~\cite{deepimageanalogy}, content style disentanglement works such as Jha et al.~\cite{jha2018disentangling}, Denton and Birodkar~\cite{denton2017unsupervised}, Bouchacourt et al.~\cite{bouchacourt}, Gabbay and Hoshen~\cite{gabbay} cannot change the structure of objects in the image. Many such works, also rely on class level supervision which we do not use.

The concept of visual analogy was explored in classical works, for example in the context of image re-targeting~\cite{michals}. Two visual signals are defined to be visually similar if all patches of one (at multiple scales) are contained in the other (\textit{completeness}), and vice versa ($\textit{coherence}$). The same concepts are relevant to our setting: our key idea is to produce a mapping in which the patch distribution of a source image is mapped to its corresponding patch distribution of a target image and vice versa. When the multi-scale distributions match, in both directions, completeness and coherence are guaranteed.

\section{Method}
\label{secmethod}

\begin{figure*}[t]
\centering
\includegraphics[width=\linewidth]{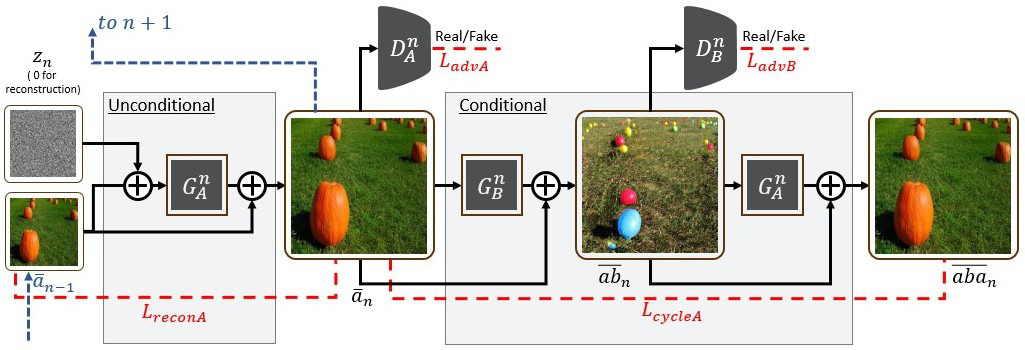}
\caption{An illustration of a single scale of our method for the image of pumpkins, $A$, and an image of balls, $B$, in the direction of $A$ to $B$ for $0 < n < K$. 
Training images are shown in Fig.~\ref{fig:images_objects}.
(Unconditional): $G^n_A$ first unconditionally generates an image of pumpkins at scale $n$, by using as input the upscaled generated pumpkin image at the previous scale, $\uparrow\overline{a}_{n-1}$, with added random noise, $z_n$. The output of $G^n_A$ is added to $\uparrow\overline{a}_{n-1}$ to generate $\overline{a}_{n}$. A patch-discriminator $D^n_A$ is used to ensure $\overline{a}_{n}$ belongs to $P^n_A$ (images of pumpkins at scale $n$). 
(Conditional): $G^n_B$ then maps $\overline{a}_n$  to $\overline{ab}_n$ by adding the minimal amount of detail to $\overline{a}_n$. A patch-disciminator $D_B^n$ ensures that $\overline{ab}_n$ belongs to $P^n_B$ (images of balls at scale $n$). $G_n^A$ is then used to map $\overline{ab}_n$ to $\overline{aba}_n$. 
(Cycle Loss): To ensure that $\overline{a}_n$ and $\overline{ab}_n$ are aligned, a cycle loss is employed between to $\overline{a}_n$ and $\overline{aba}_n$.}
\label{fig:diagram}
\end{figure*}

\begin{figure}[t]
\centering
\includegraphics[width=\linewidth]{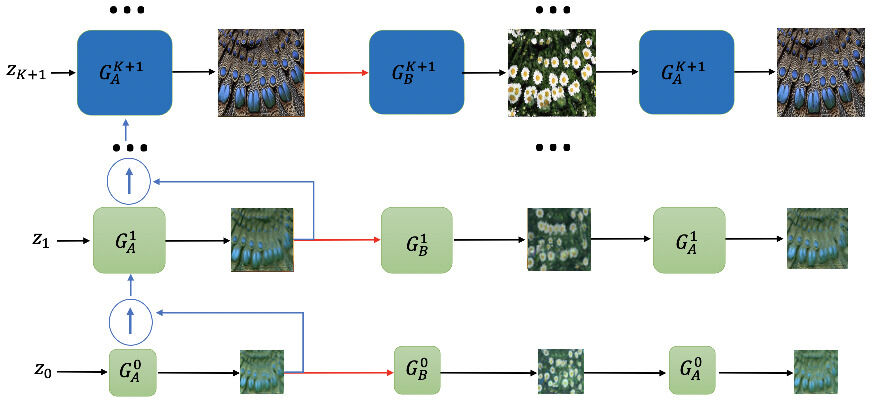}
\caption{ An illustration of the overall hierarchical structure of the generators for our method, for the image of feathers $A$, and an image of flowers $B$. We begin at scale $0$, where generators $G^0_A$ and $G^0_B$ are trained as described in Sec.~\ref{secmethod}. $\uparrow$ indicates up-scaling. $G^n_A$ and $G^n_B$ are trained for each of the scales as described in Fig.~\ref{fig:diagram}. For scale $n>K$, we switch to using non-residual training (indicated by using blue boxes for the generators).     }
\label{fig:diagram2}
\end{figure}

In unsupervised image to image translation, the learner is typically provided with images from two externally collected datasets, and considers the distributions governing these datasets to generate aligned mappings between the two distributions~\cite{CycleGAN2017,discogan,dualgan,cogan,unit,stargan,conneau2017word,zhang2017adversarial,zhang2017earth,lample2018unsupervised, DRIT_plus}.

In our setting, we consider instead the internal patch distributions of two images only, $A$ and $B$. For a given image $x$ (either $A$ or $B$), and for a particular scale $n=0,1,2,\dots,N$, let the distribution of images having the same patch distribution as $x$ at this scale be $P^n_x$. In other words, since each image is composed of many patches at different scales, we learn from the single image $x$, a distribution of images $P^n_x$, that shares the same scale-dependent patch distributions. The downsampling factor of each scale $n$ is given by $r^{N-n}$ for some $r > 1$. Scale $N$, therefore, is the original image resolution for both $A$ and $B$.  Starting at the coarsest scale $0$, and up to scale $N$, our method attempts to find a matching between each image $a_n \in P^n_A$ and an image $b_n\in P^n_B$. 

For each scale $n$, our method employs two patch GANs~\cite{CycleGAN2017, disc2} for generating samples in $P^n_A$ and $P^n_B$. Each patch-GAN consists of a fully convolutional generator $G^n_A$ and discriminator $D^n_A$ for $A$, and $G^n_B$ and $D^n_B$ for $B$. The architecture of $G^n_A$ and $D^n_A$ is outlined in Sec.~\ref{sec:arch}. 
For brevity, we describe the generation process from $A$ to $B$; the reverse generation process from $B$ to $A$ is symmetric.
An illustration of our method is provided in Fig.~\ref{fig:diagram} and Fig.~\ref{fig:diagram2}.

Our terms for both the unconditional (Figure 2 left) and conditional (Figure 2 right) generations are taken from the generative models literature \cite{singan, CycleGAN2017}. By unconditional generation, we mean the process of generating random samples of the unconditional distribution $P^n_A$. 
By conditional generation, we mean the conditional generation of $X$ given $Y$ where $X \sim P^n_B$ and $Y \sim P^n_A$.

\subsection{Generation from the unconditional distribution}
\label{sec:uncon_generation}

For a scale $n$, to map between $P^n_A$ and $P^n_B$, we first generate a sample $a \in P^n_A$ in an unconditional fashion. To do so, we follow, for the unconditional case, a similar procedure to the recently proposed SinGAN algorithm~\cite{singan}, for the first $K$ scales. Let $a_n$ be the sample $A$ down-sampled by a factor of $n$, and $z_n$ be Gaussian noise of the same shape and dimension as $a_n$. At the coarsest scale, $n=0$, the output of $G^{0}_A$ is defined as (overline  is used to denote generated images):
\begin{align}
\overline{a}_{0} = G^{0}_A(z_{0}) \label{eq:uncon1}
\end{align}
Moving to finer scales $n > 0$, the generator accepts as input both $z_{n}$ and an upscaled image of $\overline{a}_{n-1}$ to the dimension of $a_n$, denoted by $\uparrow\overline{a}_{n-1}$. 
For $n < K$, for some fixed $K$,  we opt for using residual training:
\begin{align}
\overline{a}_{n} = G^{n}_A(z_n + \uparrow\overline{a}_{n-1})  + \uparrow\overline{a}_{n-1} \label{eq:uncon2_res}
\end{align}
This teaches the network to add missing details of $\uparrow\overline{a}_{n-1}$ specific to scale $n$. For $n \geq K$, we do not use residual training:
\begin{align}
\overline{a}_{n} = G^{n}_A(z_n + \uparrow\overline{a}_{n-1}) \label{eq:uncon2_nonres}
\end{align}
$K$ is chosen as a hyperparameter which helps determine the level at which aligned mappings are produced (see Sec.~\ref{sec:alignment} for details).

\subsection{Generation from the conditional distribution}
\label{sec:cond_generation}

Given the unconditional sample $a \in P^n_A$ generated at scale $n$, we now wish to map it to its corresponding sample $b \in P^n_B$. The same generator is used for both upscaling (Eq.~\ref{eq:uncon2_res} and Eq.~\ref{eq:uncon2_nonres}) and for mapping between the domains. Note that $G^{n}_A$ learns to generate samples from $P^n_A$, regardless of whether the input is from $P^{n-1}_A$ or $P^n_B$. 
For $n < K$, we also use residual training:
\begin{align}
\overline{ab}_{n} = G^{n}_B(\overline{a}_n) + \overline{a}_n \label{eq:cond1_res}
\end{align}
For $n > K$, no residual training is applied (see Sec.~\ref{sec:alignment} for details):
\begin{align}
\overline{ab}_{n} = G^{n}_B(\overline{a}_n)\label{eq:cond1_non}
\end{align}

Equations \ref{eq:cond1_res} and \ref{eq:cond1_non} of the conditional generation are deterministic and do not require an input noise. In particular, as noted above, the input is of the same dimension as the noise matrix injected in equations \ref{eq:uncon2_res} and \ref{eq:uncon2_nonres} of the unconditional generation.
Our key observation is that for each scale, a good alignment is one where each patch in the patch distribution of $A$ is mapped to its corresponding patch in the patch distribution of $B$ (and vice versa), and where the relative position of objects is preserved. 

The corresponding patch is learned to be the patch in $P^n_B$ such that $G^n_A$ can reconstruct the original input patch from it with minimum error. 
This enforces a one-to-one correspondence between the patches of A and the patches of B at a given scale. 

The goal of this mapping is, therefore, that: (1) at the coarsest scale $n=0$, when mapping an image $a \in P^0_A$ to $b \in P^0_B$, $G^0_B$ should map global structure of $a$ to global structure of $b$. For example, the ground at the bottom of $a$ should be mapped to grass at the bottom of $b$, and daylight in $a$ may be mapped to darker light in $b$. (2) at finer scales $n > 0$, $G^n_B$ generates finer details in $P^n_B$, and so beyond the global, $G^n_B$ should map finer details, such as style or texture in $a$ to such details in $b$. This goal is achieved through the various loss terms employed by our method.

\subsection{Loss Terms}

\paragraph*{Adversarial Loss}
To ensure that $\overline{a}_n$ belongs to $P^n_A$ and $\overline{ab}_n$ belongs to $P^n_B$, we employ adversarial patch-GAN loss using a markovian discriminator $D^A_n$~\cite{CycleGAN2017, disc2}. $D^A_n$ is fully convolutional and uses the same architecture as $G^A_n$ to produce a map of the same dimension of $\overline{a}_n$. $D^A_n$ maps each (overlapping) patch of its input as real or fake. In particular, $G^A_n$ attempts to fool $D^A_n$ into classifying $\overline{a}_n$'s patches as real, while $D^A_n$ attempts to classify $\overline{a}_n$'s patches as fake and $a_n$'s patches as real. We use the following WGAN-GP~\cite{wgan-gp} loss for $G^A_n$ and $D^A_n$:
\begin{align}
\mathcal{L}^1_{adv}(D^A_n, G^A_n) = \E(D^A_n(a_n)) - \E(D^A_n(\overline{a}_n)) - GP^n_A(a_n, \overline{a}_n)
\end{align}

where $\E$ is mean over $D^A_n$'s output and the gradient penalty of WGAN-GP~\cite{wgan-gp} is defined as:
\begin{align}
GP^n_A (a, b) = - \lambda \E [(||\nabla_{c} D^A_n(c) -1||_2)^2]
\end{align}
where $c = \epsilon a + (1-\epsilon)b$ for $\epsilon$ sampled uniformly between $0$ and $1$ and $\lambda$ is an hyperparameter.

The fake image $\overline{a}_n$ is randomly generated by $G_n^A$, using equations \ref{eq:uncon1}, \ref{eq:uncon2_res} and \ref{eq:uncon2_nonres} by sampling $z_n$ as a Gaussian noise. Note that the losses are defined for a batch size of one, as used in our experiments.

Similarly, 
$G^B_n$ attempts to fool $D^B_n$ into classifying $\overline{ab}_n$'s patches as real, while $D^B_n$ attempts to classify $\overline{ab}_n$'s patches as fake and $b_n$'s patches as real:

\begin{align}
\mathcal{L}^2_{adv}(D^B_n, G^B_n) = \E(D^B_n(b_n)) - \E(D^B_n(\overline{ab}_n)) - GP^n_B(b_n, \overline{ab}_n)
\end{align}

We also consider the analogous losses in the direction of $B$ to $A$:

\begin{align}
\mathcal{L}^1_{adv}(D^B_n, G^B_n) &= \E(D^B_n(b_n)) - \E(D^B_n(\overline{b}_n)) - GP^n_B(b_n, \overline{b}_n) \\ \label{eq:uncod_eq1b}
\mathcal{L}^2_{adv}(D^A_n, G^A_n) &= \E(D^A_n(a_n))  - \E(D^A_n(\overline{ba}_n)) - GP^n_A(a_n, \overline{ba}_n)
\end{align}
We sum these four loss terms to define:
\begin{align}
  \mathcal{L}_{adv_n} & = \mathcal{L}^1_{adv}(D^A_n, G^A_n) + \mathcal{L}^1_{adv}(D^B_n, G^B_n)  \nonumber \\ &+ \mathcal{L}^2_{adv}(D^A_n, G^A_n) + \mathcal{L}^2_{adv}(D^B_n, G^B_n)
\end{align}

\paragraph*{Reconstruction Loss}
\label{sec:reconstruction_loss}
For $n>0$, when no noise is used, we would like the generator $G^n_A$ to reconstruct $a_n$ and  $G^n_B$ to reconstruct $b_n$.  Specifically:
\begin{align}
\mathcal{L}^A_{recon_n} &= ||G^{n}_A(\uparrow\overline{a}_{n-1}) - a_n||_2 \\
\mathcal{L}^B_{recon_n} &= ||G^{n}_B(\uparrow\overline{b}_{n-1}) - b_n||_2
\end{align}

When $n=0$, we employ a fixed random noise $z_0 = z_a^*$ or $z_0 = z_b^*$ and apply the losses: 
$\mathcal{L}^A_{recon_0} = ||G^{0}_A(z_a^*) - a_0||_2$, $\mathcal{L}^B_{recon_0} = ||G^{0}_B(z_b^*) - b_0||_2$. 
We define:
\begin{align}
\mathcal{L}_{recon_n} &= \mathcal{L}^A_{recon_n} + \mathcal{L}^B_{recon_n} \label{eq:recon_loss}
\end{align}

The reconstruction loss is used to control $\sigma_n$, the standard deviation of the Gaussian noise $z_n$, which indicates the level of detail required at each scale. In particular, $\sigma_n$ is taken to be $||\uparrow\overline{a}_{n-1} - a_n||_2$. In addition, it acts as the identity loss used in CycleGAN~\cite{CycleGAN2017}. Without it, $G_n^A$ and $G_n^B$ 
can tint input images, as shown in the ablation study of Sec.~\ref{sec:ablation}.  

\paragraph*{Cycle Loss.}
\label{sec:cycle_loss}
To encourage the mapping between the two patch distribution to be aligned, we employ a cycle loss~\cite{CycleGAN2017, dualgan, discogan} for $n=0,1,..,K-1$. For $n \geq K$, it is not applied.
The mapping back from $\overline{ab}_{n}$ to $P^n_A$ and from $\overline{ba}_{n}$ to $P^n_B$ is given as $\overline{bab}_{n} = G^{n}_B(\overline{ba}_n)$ and $\overline{bab}_{n} = G^{n}_B(\overline{ba}_n)$. The cycle loss is given by:

\begin{align}
\mathcal{L}_{cycle_n} & = ||\overline{a}_n - \overline{aba}_n||_2  + ||\overline{b}_n - \overline{bab}_n||_2 \label{eq:cycle_loss}
\end{align}

\noindent The overall loss at scale $n$, for two hyper-parameters $\lambda_{cycle},\lambda_{recon}>0$, is:
\begin{align}
  \mathcal{L}_n  &= \min\limits_{G^A_n, G^B_n} \max\limits_{D^A_n, D^B_n} \mathcal{L}_{adv_n} + \lambda_{recon}\mathcal{L}_{recon_n} + \lambda_{cycle}\mathcal{L}_{cycle_n} \label{eq:full_loss}
\end{align}

\subsection{Training}
\label{sec:alignment}

Our algorithm trains the networks of each scale one at a time, keeping the networks of the previous scales fixed. In particular, after a set number of epochs (typically $10,000$), we stop training the networks at scale ${n-1}$ and move to scale ${n}$. When moving to scale ${n}$, the weights of the networks at this scale are initialized with the weights of the networks at the previous scale $n-1$. As shown in Sec.~\ref{sec:ablation}, such initialization is required to generate aligned mappings.

Our method enables a control over the type of alignment produced. This is in part through the choice of $K$, whose effect is demonstrated in Sec.~\ref{sec:ablation}.
For $n < K$, we use residual training for both unconditional and conditional generation (Eq.~\ref{eq:uncon2_res} and Eq.~\ref{eq:cond1_res}). 
For the unconditional generation, this teaches the network to add missing details of $\uparrow\overline{a}_{n-1}$ specific to scale $n$. 
For the conditional generation setting, $G^{n}_B$ learns to add the minimal amount of detail required for each patch of $a_n$, so as to generate an image belonging to $P^{n}_B$ and to satisfy the cycle loss. We find that using this type of residual training is essential for generating aligned solutions.

From scale $n \geq K$ onward, the generator of the target domain does not need to maintain the exact details of the input image, which allows it the freedom to be faithful to the details of the target domain, thus improving the generation quality. This allows for a level of control in the alignment produced. For $n \geq K$, we, therefore, opt for a non-residual architecture, where $G^{n}_A$ simply generates an image $\overline{ab}_n$, conditioned on $\overline{a}_n$.

Note that there are other factors that determine the type of alignment such as the total number of scales used, the individual image resolutions used at different scales and the receptive field of the generators and discriminators. See Sec.~\ref{sec:arch} for additional details.

\paragraph*{Interaction between scales}
We observe that using a conditional generation pipeline that is directly conditioned on the previous scale does not produce aligned solutions. 
In particular, conditioning $G^{n}_B$ on both $\overline{a}_n$ and $\uparrow\overline{ab}_{n-1}$ (the mapping of $\overline{a}_{n-1}$ at the previous scale), the algorithm learns to `cheat' by embedding details in $\overline{a}_n$ in $\overline{ab}_n$, which still allow it to satisfy the cycle loss, while not producing aligned solutions (see ablation Sec.~\ref{sec:ablation}). 
Instead, we incorporate the interaction indirectly, using the same identical network for the conditional and unconditional generation. Since networks at different scales directly interact in the unconditional pipeline, this induces an indirect interaction for the conditional one.

Our experiments show that inference at the final scale is affected by the training of previous ones, and hence that the final scale captures the indirect interaction between networks at previous scales. Sec.\ref{sec:ablation} demonstrates this by showing that omitting the cycle loss in all but scale $K-1$, results in bad alignment.
In addition, Sec.\ref{sec:ablation} demonstrates that using two separate networks for the conditional and unconditional generation results in unaligned solutions. This is because the indirect interaction between scales, as described above, is no longer possible.

\subsection{Inference and Refinement}
\label{sec:inference}

At inference time, we change $A$'s resolution to that of a chosen scale $S$. We then upsample (super-resolve) it using the $A$'s unconditional generation pipeline up to the final scale $N$. Finally, we map the resulting image to the target domain.

More precisely, we replace $\overline{a}_S$ with a resized image of $A$ ($= a_S$). We refer to $S$ as the "injection scale", and the act of replacing $\overline{a}_S$ with $a_S$ as "injecting". 
We then upscale  $\overline{a}_S$ as in Eq.~\ref{eq:uncon2_res} and Eq.~\ref{eq:uncon2_nonres} to generate $a^*_N \in P^N_A$, where the $^*$ notation is used to differentiate from $a_N$. Finally we map $a^*_N$ to  $\overline{ab}_N$ using Eq.~\ref{eq:cond1_non}.

$S$ is a hyperparameter, typically chosen to be $N-2$. 
This inference procedure could also be applied for a randomly generated unconditional sample $\overline{a}_S$ instead of $a_S$, resulting in random aligned pairs that are not necessary aligned to the input. An analysis of the effect of the choice of $S$ as well as alternative inference procedures are given in Sec.~\ref{sec:inference_ablation}.

For simplicity of notation, we write $P^N_A$ as $P_A$ and the mapping from $A$ ($ = a_S$) to $P^N_B$ as $ab$. We write $\overline{a}_N$ as $\overline{a}$ and $\overline{ab}_N$ as $\overline{ab}$. Similar annotation is used for $B$. 
To improve the quality of $\overline{ab}_N$, we found it useful to use an additional refinement procedure in some cases. 
We train a separate `refinement' network, which contains only the unconditional generation modules (Sec.~\ref{sec:uncon_generation}) and losses (Eq.~\ref{eq:uncod_eq1b} and Eq.~\ref{eq:recon_loss}) for image $b$. 
We then insert $\overline{ab}_N$ to this network at a fine scale. This adds fine details to the image, while not affecting the overall alignment.

\subsection{Architecture and Implementation details}
\label{sec:arch}

The generators $G^n_A$ and $G^n_B$, as well as the discriminators $D_n^A$ and $D_n^B$, each consist of five convolutional blocks. Each block consists of a $3\times 3$ convolutional layer and maintains the spatial resolution of the input using padding. After the convolutional layer, batch norm and a LeakyReLU (with slope of $0.2$) activation are used. This results in a fixed effective receptive field of $11\times11$ for all generators and discriminators at each scale. As the input $a_n$ increases in size for each scale, our algorithm starts by matching large patches between $P^n_A$ and $P^n_B$ (global structure and alignment of objects) and finishes at matching patches of smaller size. For the last convolutional block no batch norm is used and LeakyReLU is replaced with $tanh$. An adam optimizer is used with learning rate of $0.0005$ and parameters $\beta_1 = 0.5$ and $\beta_2 = 0.999$. 
In Eq.~9, $\lambda_{recon} = 1.0$  and $\lambda_{cycle} = 10.0$ for all experiments.

The hyperparameters $K$, $r$ and $N$ (see Sec.~3) are set as follows: $r$ is chosen to be $0.75$ in all experiments; $N$ is chosen such that the maximal image size at the finest scale is $220px$ and the minimal image size at the coarsest scale is $18px$. We used $K=N-1$ for style and texture transfer experiments and $K=N-2$ or $K=N-1$ for the other experiments.

\section{Results}

We evaluate our method qualitatively and quantitatively on a variety of image pairs collected from the Berkeley Segmentation Database~\cite{bsd}, Places~\cite{places} and the web. We consider pairs of images from different natural scenes, such as a mountain with snow and that of a pyramid on sand. We also considered objects with different local shape and structure, such as an image of hot air balloons and that of birds, to evaluate the ability of our method to adapt to the local structure of objects. While the main application of our method is that of structural alignment, we also evaluate our method on the conditional generation tasks of guided image synthesis, style and texture transfer, text translation and video translation. Lastly, an ablation analysis is performed to illustrate the effectiveness of the different components of our method.

\subsection{Structural Alignment}

\begin{figure*}
  \centering
  \begin{tabular}{c@{~}c:c@{~}c@{~}c@{~}c@{~}c}
  & Input & \textbf{Ours} & DIA~\cite{deepimageanalogy} & SinGAN~\cite{singan} & Cycle$^*$~\cite{CycleGAN2017} & Style~\cite{Gatys_2016_CVPR} \\
  
\centering{\begin{small}\begin{turn}{90}\;\;\;\;\;\; \hspace{-20mm} Pumpkins2Balls \end{turn}\end{small}}&
\includegraphics[width=0.140\linewidth, clip]
{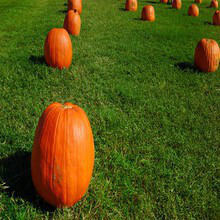}&
\includegraphics[width=0.140\linewidth,clip] {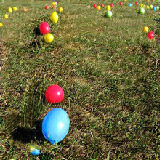}&
\includegraphics[width=0.140\linewidth,clip] 
{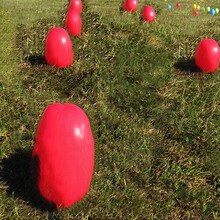}&
\includegraphics[width=0.140\linewidth,clip] {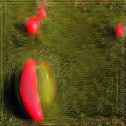}&
\includegraphics[width=0.140\linewidth,clip] {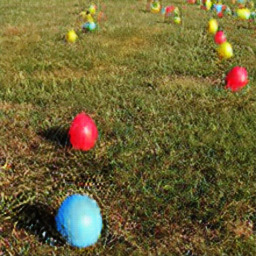}& 
\includegraphics[width=0.140\linewidth,clip] {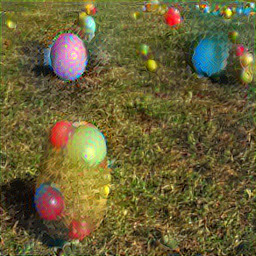}\\

&
\includegraphics[width=0.140\linewidth, clip]
{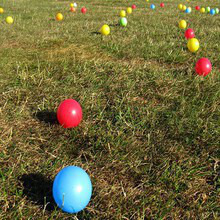}&
\includegraphics[width=0.140\linewidth,clip] {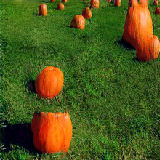}&
\includegraphics[width=0.140\linewidth,clip] 
{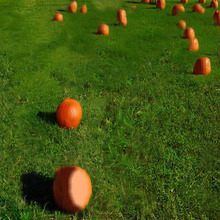}&
\includegraphics[width=0.140\linewidth,clip] {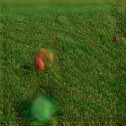}&
\includegraphics[width=0.140\linewidth,clip] {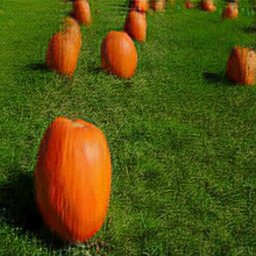}& 
\includegraphics[width=0.140\linewidth,clip] {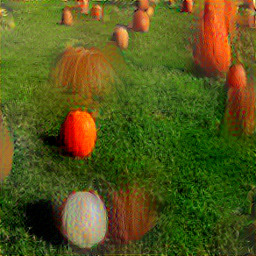}\\

\centering{\begin{small}\begin{turn}{90}\;\;\;\;\;\; \hspace{-24mm} Footprints2SnowSteps \end{turn}\end{small}}&
\includegraphics[width=0.140\linewidth, clip]
{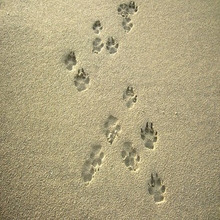}&
\includegraphics[width=0.140\linewidth,clip] {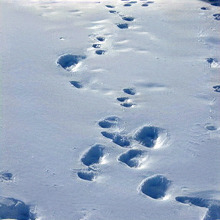}&
\includegraphics[width=0.140\linewidth,clip] 
{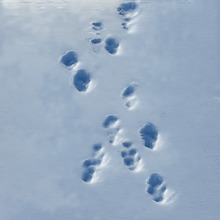}&
\includegraphics[width=0.140\linewidth,clip] {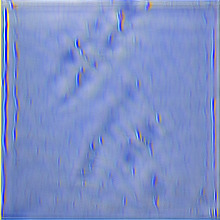}&
\includegraphics[width=0.140\linewidth,clip] {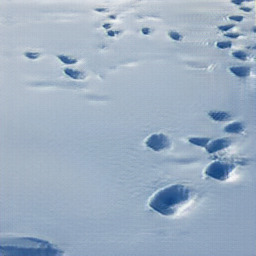}& 
\includegraphics[width=0.140\linewidth,clip] {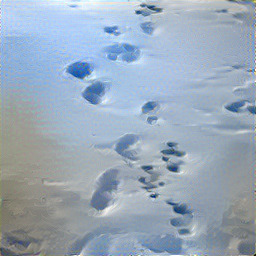}\\

&
\includegraphics[width=0.140\linewidth, clip]
{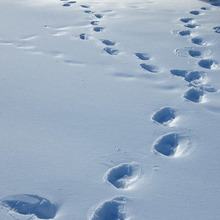}&
\includegraphics[width=0.140\linewidth,clip] {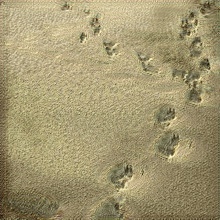}&
\includegraphics[width=0.140\linewidth,clip] 
{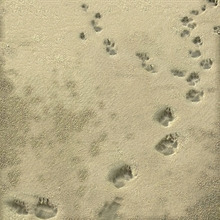}&
\includegraphics[width=0.140\linewidth,clip] {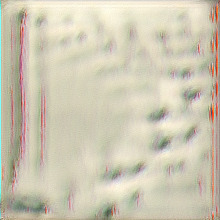}&
\includegraphics[width=0.140\linewidth,clip] {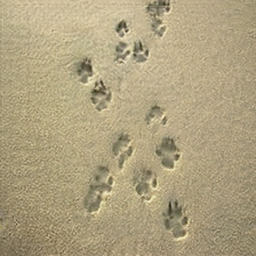}& 
\includegraphics[width=0.140\linewidth,clip] {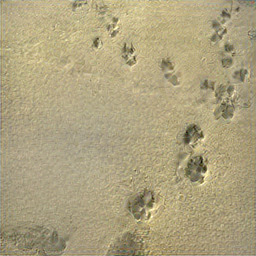} \\
  
\centering{\begin{small}\begin{turn}{90}\;\;\;\;\;\; \hspace{-18mm} Balls2Marbles \end{turn}\end{small}}&
\includegraphics[width=0.140\linewidth, clip]
{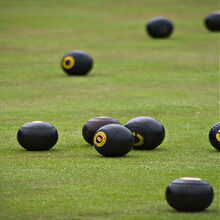}&
\includegraphics[width=0.140\linewidth,clip] {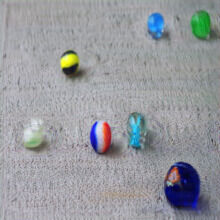}&
\includegraphics[width=0.140\linewidth,clip] 
{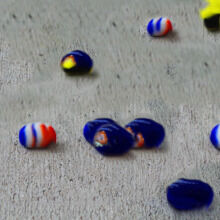}&
\includegraphics[width=0.140\linewidth,clip] {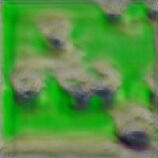}&
\includegraphics[width=0.140\linewidth,clip] {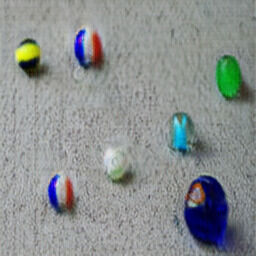}& 
\includegraphics[width=0.140\linewidth,clip] {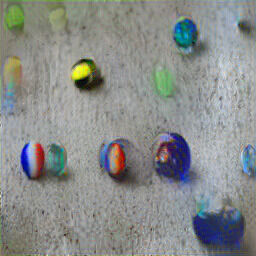}\\ 

&
\includegraphics[width=0.140\linewidth, clip]
{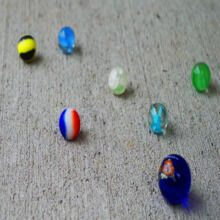}&
\includegraphics[width=0.140\linewidth,clip] {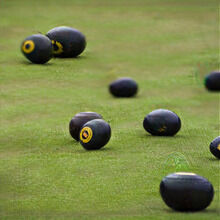}&
\includegraphics[width=0.140\linewidth,clip] 
{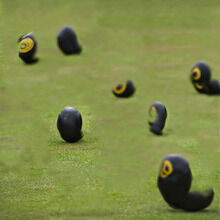}&
\includegraphics[width=0.140\linewidth,clip] {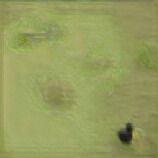}&
\includegraphics[width=0.140\linewidth,clip] {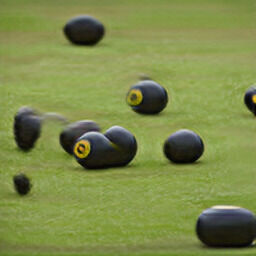}& 
\includegraphics[width=0.140\linewidth,clip] {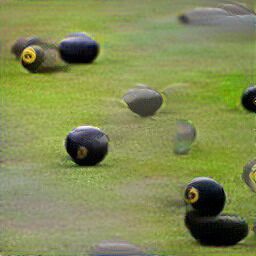}\\ 

\centering{\begin{small}\begin{turn}{90}\;\;\;\;\;\; \hspace{-16mm} Plant2Logs \end{turn}\end{small}}&
\includegraphics[width=0.140\linewidth, clip]
{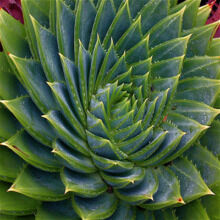}&
\includegraphics[width=0.140\linewidth,clip] {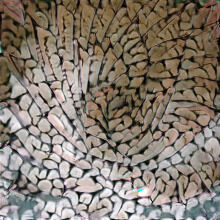}&
\includegraphics[width=0.140\linewidth,clip] 
{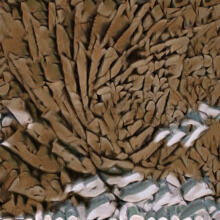}&
\includegraphics[width=0.140\linewidth,clip] {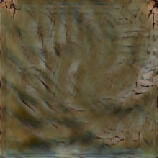}&
\includegraphics[width=0.140\linewidth,clip] {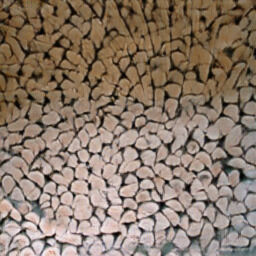}& 
\includegraphics[width=0.140\linewidth,clip] {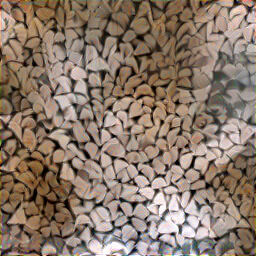}\\ 

&
\includegraphics[width=0.140\linewidth, clip]
{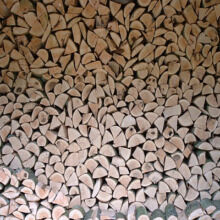}&
\includegraphics[width=0.140\linewidth,clip] {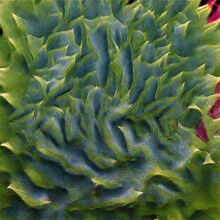}&
\includegraphics[width=0.140\linewidth,clip] 
{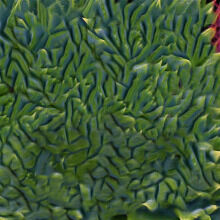}&
\includegraphics[width=0.140\linewidth,clip] {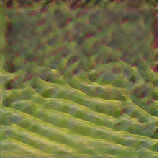}&
\includegraphics[width=0.140\linewidth,clip] {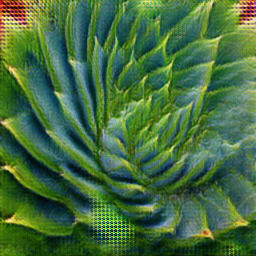}& 
\includegraphics[width=0.140\linewidth,clip] {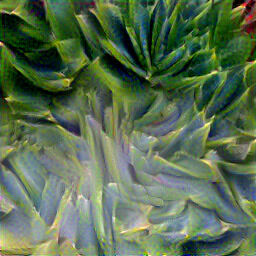}\\ 
\end{tabular}
\caption{Structural alignment for a collection of images containing objects of different local structure. We consider the mapping of our method and baseline ones. For every two rows, the first row is the translation from $A$  to $ab$ and the second is from $B$ to $ba$. 
Cycle$^*$ stands for CycleGAN applied on randomly generated images using pretrained SinGAN models for $A$ and $B$. 
}
\label{fig:images_objects}
\end{figure*}

\begin{figure*}
  \centering
  \begin{tabular}{c@{~}c:c@{~}c@{~}c@{~}c@{~}c}
  & Input & \textbf{Ours} & DIA~\cite{deepimageanalogy} & SinGAN~\cite{singan} & Cycle$^*$~\cite{CycleGAN2017} & Style~\cite{Gatys_2016_CVPR} \\
  
\centering{\begin{small}\begin{turn}{90}\;\;\;\;\;\; \hspace{-21mm} Mountains2Pyramid \end{turn}\end{small}}&
\includegraphics[width=0.140\linewidth, clip]
{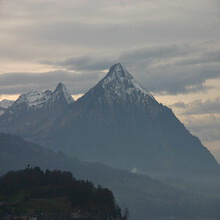}&
\includegraphics[width=0.140\linewidth,clip] {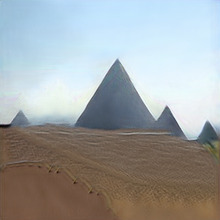}&
\includegraphics[width=0.140\linewidth,clip] 
{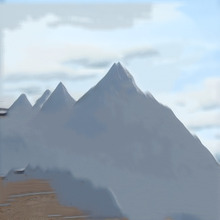}&
\includegraphics[width=0.140\linewidth,clip] {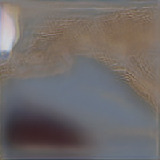}&
\includegraphics[width=0.140\linewidth,clip] {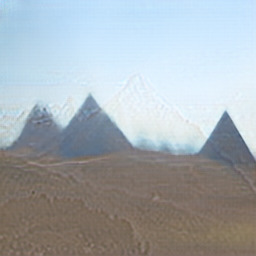}& 
\includegraphics[width=0.140\linewidth,clip] {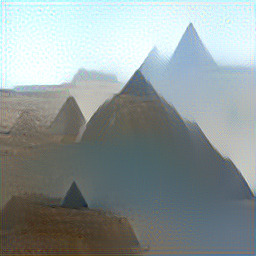}\\

&
\includegraphics[width=0.140\linewidth, clip]
{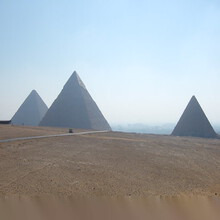}&
\includegraphics[width=0.140\linewidth,clip] {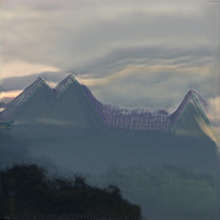}&
\includegraphics[width=0.140\linewidth,clip] 
{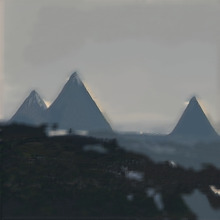}&
\includegraphics[width=0.140\linewidth,clip] {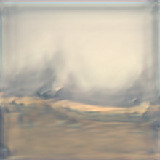}&
\includegraphics[width=0.140\linewidth,clip] {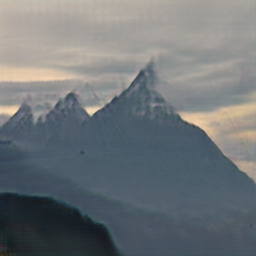}& 
\includegraphics[width=0.140\linewidth,clip] {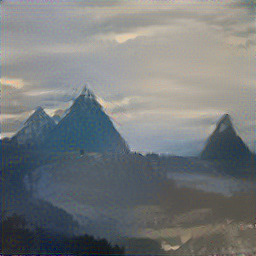} \\

\centering{\begin{small}\begin{turn}{90}\;\;\;\;\;\; \hspace{-16mm} Ducks2Orcas \end{turn}\end{small}}&
\includegraphics[width=0.140\linewidth, clip]
{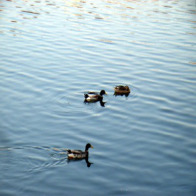}&
\includegraphics[width=0.140\linewidth,clip] {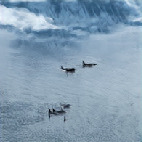}&
\includegraphics[width=0.140\linewidth,clip] 
{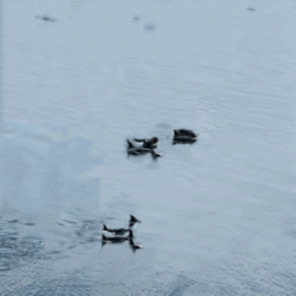}&
\includegraphics[width=0.140\linewidth,clip] {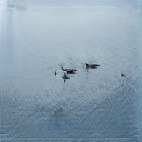}&
\includegraphics[width=0.140\linewidth,clip] {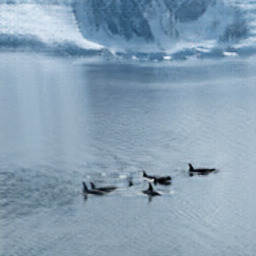}& 
\includegraphics[width=0.140\linewidth,clip] {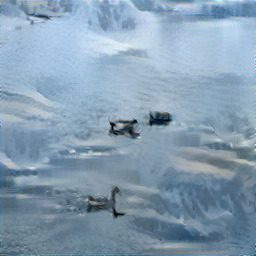}\\

&
\includegraphics[width=0.140\linewidth, clip]
{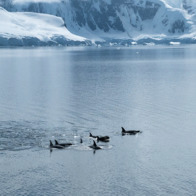}&
\includegraphics[width=0.140\linewidth,clip] {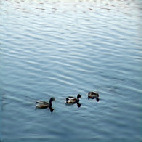}&
\includegraphics[width=0.140\linewidth,clip] 
{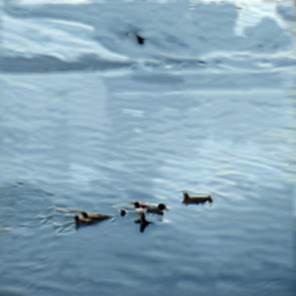}&
\includegraphics[width=0.140\linewidth,clip] {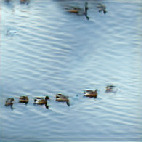}&
\includegraphics[width=0.140\linewidth,clip] {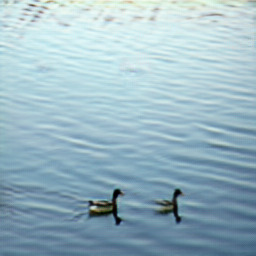}& 
\includegraphics[width=0.140\linewidth,clip] {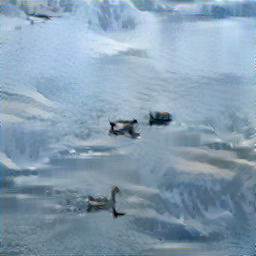}\\

\end{tabular}
\caption{Structural alignment for a collection of images depicting natural scenes. We consider the same setting as in Fig~.\ref{fig:images_objects}. 
}
\label{fig:images_scenes}
\end{figure*}

\paragraph*{Qualitative evaluation}
\label{sec:visual_results}
In Fig.~\ref{fig:images_objects} and Fig.~\ref{fig:images_scenes}, we illustrate the structural alignment produced by our method for a variety of image pairs. Additional results are provided in the appendix in Figs.~\ref{fig:additional2}, \ref{fig:additional3}, \ref{fig:additional4} and \ref{fig:additional5}. We evaluate our method against four baselines. First, we consider Deep Image Analogy (DIA)~\cite{deepimageanalogy}, since it is the closest work to ours. A second baseline is that of SinGAN~\cite{singan}. Since SinGAN is aimed at unconditional image generation from a single image, we first train a SinGAN model for image $B$. We then insert a downsampled version of image $A$ at a coarse scale of $B$'s pretrained model (chosen to be 2, see Sec.~\ref{sec:baselines}) and use the image output from the last scale. Thirdly, we train two SinGAN models for $A$ and $B$ and randomly generate $1000$ images from the internal distribution of each image. We then train a CycleGAN model~\cite{CycleGAN2017} on these images. Lastly, we consider style transfer from $A$ to $B$ using the method of Gatys et al.~\cite{Gatys_2016_CVPR}. Note that unlike our method, DIA~\cite{deepimageanalogy} and Gatys et al.~\cite{Gatys_2016_CVPR} use an additional level of supervision in the form of deep features of a pretrained VGG~\cite{vgg} network trained on Imagenet.  

In Fig.~\ref{fig:images_objects}, we consider a variety of image pairs involving local objects of different shapes and structure. Unlike the baseline methods, our method can adapt to the structure and shape of objects in the target image. For example, when translating from an image of pumpkins to an image of balls of different sizes, our method is able to replace the large pumpkin (bottom left) with two balls in the correct position and shape, since there is no ball as large as the pumpkin. DIA, on the other hand, is unable to replace the shape of the pumpkin, producing an unrealistic image. SinGAN is unable to create an image of good quality, since it was trained on an image with a different patch distribution. Our third baseline (CycleGAN trained on images generated using two SinGANs) produces a realistic looking image of balls, but is unable to produce aligned solutions and seems to suffer from mode collapse, returning an image that is rather similar to the original one. 

Our method generates analogies that keep the correct scales of objects, as they appear in the source image.  
Consider, for example, the image pair of pumpkins and balls in Fig~.\ref{fig:images_objects}. The largest ball is smaller than the largest pumpkin, so
the pumpkin is mapped to two balls instead of one. In the other direction, since small pumpkins that correspond in size to small balls exist, a single ball is mapped to a single pumpkin. We argue that this mapping is desirable. In DIA~\cite{deepimageanalogy}, the largest pumpkin is mapped to a large ball, but this no longer looks realistic. We, therefore, argue that our multi scale approach correctly handles image pairs exhibiting significantly different scales of objects. 

By producing a "structural analogy", we produce a mapping where structure is preserved at all scales. For example generated samples do not display bird-shaped balloon strips, nor do they position crowds in the sky. Further, in guided image synthesis (see Fig.~\ref{fig:sketchtoimage}, we preserve the structure of a simple sketch that does not contain any texture.

When translating from footprints of a dog in the sand to those of a human in the snow, our method is able to create realistically looking steps in the correct position, while preserving the perspective warp the target snow image displays. DIA and Gatys et al.~\cite{Gatys_2016_CVPR} are unable to change the shape of the footprints, while CycleGAN produces an unaligned solution. 
Similarly, our method is also able to align many small objects of different geometries, as depicted in the last two row pairs of the figure. 

Fig.~\ref{fig:images_scenes} illustrates this translation between different natural scenes. Here, again, we see our method succeed in finding plausible analogies (e.g. mountains to mountains, or animals to animals) and to place them correctly, while the baselines exhibit failure modes similar to those discussed beforehand. 

Note that our method can handle image pairs of different perspectives. For instance, Fig.~\ref{fig:additional2} of the appendix depicts an image pair of flowers and cows. In addition, Fig.~\ref{fig:teaser}  and the first example of Fig.~\ref{fig:additional2} of the appendix considers an image pair with different vanishing points. In both cases, our method produces aligned results, as the difference in perspective does not change the scale of the individual objects.

In Fig.~\ref{fig:random_samples}, we consider images of hot air balloons ($A$) and of birds ($B$). We generated random samples $\overline{a} \in P_A$ and their analogous solutions $\overline{ab}$, as described in Sec.~\ref{sec:inference} (and similarly for $B$). While small balloons are mapped to small birds, balloons of large size are mapped to a collection of birds, since no bird of matching size is present in $P_B$. Similarly, a flock of birds is mapped to a large balloon in the reverse direction.

To understand the generality of the approach on completely random samples, we chose two domains from the Places dataset~\cite{places}, and then randomly selected an image from each domain. 
As can be seen in the appendix Fig~\ref{fig:rand}, the generated samples are of good quality, and some alignment is present. 
Note that when translating between images that are extremely different, the expected outcome is not always clear. For example, when translating between an image of a face and an image of a street. Hence, our method is expected to work on pairs where it is possible to understand what the desired mapping should be.

\paragraph*{Quantitative evaluation}
\label{sec:quantitative}
To evaluate the structural alignment produced by our method, we measure the following: (C1) The realism of generated samples $ab$ under the distribution of $P_B$. 
(C2) Structural alignment of image $ab$ to $A$. 

For (C1) we use the SIFID measure introduced by SinGAN~\cite{singan}. SIFID is an extension of the popular FID metric~\cite{fid} for a single image, used to evaluate how well the generated samples capture the internal statistics of the single image.

To further evaluate (C1), we conducted a user study. Our study is comprised of $50$ users and $20$ image pairs. 
All image pairs used for the user study are shown in Figures \ref{fig:images_objects}, \ref{fig:images_scenes} and in the appendix Figs~\ref{fig:additional2}, \ref{fig:additional3}, \ref{fig:additional4} and \ref{fig:additional5}.
For each image pair $A$ and $B$, the real image $A$ is shown first, followed by our and the baseline methods' translations $ab$ at random. The user is asked to rank how real each generated image looks,compared to the original image, from a scale of $1$ to $5$. 
In Tab.~\ref{tab:userstudy_sifid}, we report both the average SIFID and a mean opinion score for $50$ samples $ab$ (mapped from $A$), as described in Sec.~\ref{sec:inference}. 

To evaluate (C2), we have also conducted a user study. For each each image pair, we train a SinGAN model on $A$ and randomly generate three other samples from $P_A$: $\overline{a}_1$ ,$\overline{a}_2$ and $\overline{a}_3$. For our method and each of the baseline methods at random, the user is shown the image $ab$ and is asked to select which of $\overline{a}_1$, $\overline{a}_2$ and $\overline{a}_3$ and $A$ (shown at random), is the correct source image. The better the alignment, the easier this task is for the user. The percentage of correct answers is reported in Tab.~\ref{tab:userstudy_sifid}.  Fig.~\ref{fig:c2} of the appendix gives an illustration of how the (C2) user study was conducted when presented to the user.

For the task of structural alignment, our method is superior to all baselines in realism, and fall slight short compared to DIA in alignment. As shown visually in Fig.~\ref{fig:images_objects} and Fig.~\ref{fig:images_scenes}, DIA is unable to change the shape of translated images and so, while the user can easily identify the correct mapping, its realism scores are significantly worse then ours.

\begin{table*}
\begin{center}
\caption{Measuring the realism and structural alignment of generated samples $ab$ for our method and for baseline methods. Row (1): average SIFID values for $50$ generated images $ab$ (lower value is better). 
Row (2): User study for (C1: realism), the realism of generated samples $ab$, measured as a mean opinion score on a scale of $1$  to $5$ (higher is better). Row (3): User study for (C2: alignment), structural alignment of $ab$ to $A$, measured as the percentage of correct answers (higher is better). Rows (4-6): As for rows (1-3), but for the guided image synthesis task.
Rows (7-8): Two object classes. Row (7) measures realism as for row 2. Row (8) measures ``Natural Mapping'': Are objects of similar shape and size mapped to each other? Both measured as a mean opinion score on a scale of $1$  to $5$ (higher is better)}.
\begin{tabular}{llccccc}
\toprule
Measure & Task &\textbf{Ours}  & DIA~\cite{deepimageanalogy}  & SinGAN~\cite{singan} & Cycle$^*$~\cite{CycleGAN2017} &  Style~\cite{Gatys_2016_CVPR}  \\  
\midrule
SIFID $\downarrow$ & Structural alignment & $0.097$ & $0.723$ & $1.455$ & $0.099$ & $0.103$ \\
Realism $\uparrow$ & Structural alignment &  $3.72$ & $2.00$ & $1.57$ & $3.52$ & $1.62$ \\
Alignment $\uparrow$ & Structural alignment & $83.4\%$ & $85.0\%$ & $55.2\%$ & $43.3\%$ & $78.3\%$ \\
\midrule
SIFID $\downarrow$ & Guided synthesis & $0.191$ & $0.716$ & $1.573$ & $0.235$ & $0.208$ \\
Realism $\uparrow$ & Guided synthesis &  $4.09$ & $1.83$ & $1.47$ & $4.16$ & $1.83$ \\
Alignment $\uparrow$ & Guided synthesis & $96.3\%$ & $100.0\%$ & $77.8\%$ & $37.0\%$ & $74.1\%$ \\
\midrule
Realism $\uparrow$ & Two Object Classes &  $4.17$ & $2.52$ & $1.13$ & $1.63$ & $1.96$ \\
Natural Mapping $\uparrow$ & Two Object Classes & $4.21$ & $3.03$ & $1.31$ & $1.63$ & $2.54$
\\
\bottomrule
\end{tabular}
\label{tab:userstudy_sifid} 
\end{center}
\vspace{-0.3cm}
\end{table*}

\paragraph*{Two object classes}

We also consider the case where images depict two objects classes. Fig.~\ref{fig:multi_result} illustrates our translation for the case where A is a picture of two sea horses and a starfish and B is a picture of a flower and two tall bamboos. As expected, the starfish is mapped to the flower (and vice versa) as both have similar structure and so are the seahorses and the bamboos, demonstrating consideration of structure in the analogy.

To further demonstrate this, we constructed additional synthetic samples, which depict two or more objects classes that vary in shape, size, location, and repetition such that each pair has one mapping which is clearly more natural. We conducted two user studies following the setting of C1 above: (1) Realism (as for C1 above). (2) ``Natural Mapping'': Are objects of similar shape and size mapped to each other? Mean Opinion Scores are provided in Tab.~\ref{tab:userstudy_sifid}. As can be seen, our method is superior to baselines on both user studies. This further illustrate the ability of our method to produce ``structural analogies'', as objects of similar shape and size mapped to each other

\subsection{Additional applications}
\label{sec:addapps}

\begin{figure*}
  \centering
  \begin{tabular}{c@{~}c@{~}c:c@{~}c@{~}c@{~}c@{~}c@{~}c}
  & \scriptsize{Sketch ($A$)} & \scriptsize{Input ($B$)} & \scriptsize{\textbf{Ours}} &
  \scriptsize{DIA~\cite{deepimageanalogy}} &
  \scriptsize{SinGAN~\cite{singan}}  &  \scriptsize{Cycle$^*$~\cite{CycleGAN2017}} & \scriptsize{Style~\cite{Gatys_2016_CVPR}} \\

\centering{\begin{small}\begin{turn}{90}\;\;\;\;\;\; \hspace{-5mm} \end{turn}\end{small}}&
\includegraphics[width=0.12\linewidth, clip]
{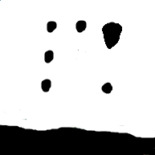}&
\includegraphics[width=0.12\linewidth,clip] {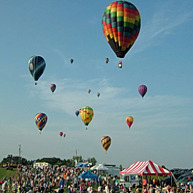}&
\includegraphics[width=0.12\linewidth,clip] 
{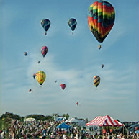}&
\includegraphics[width=0.12\linewidth,clip] {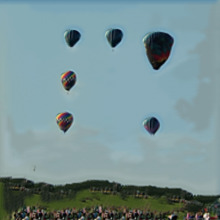}&
\includegraphics[width=0.12\linewidth,clip] {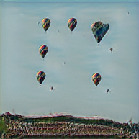}&
\includegraphics[width=0.12\linewidth,clip] {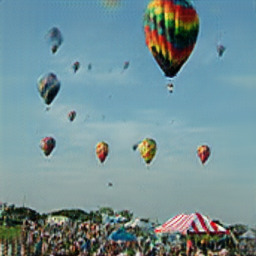}& 
\includegraphics[width=0.12\linewidth,clip] {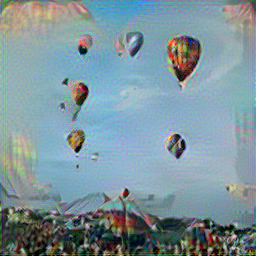}\\

\centering{\begin{small}\begin{turn}{90}\;\;\;\;\;\; \hspace{-5mm} \end{turn}\end{small}}&
\includegraphics[width=0.12\linewidth, clip]
{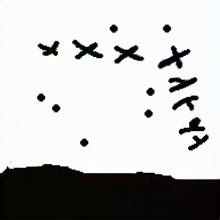}&
\includegraphics[width=0.12\linewidth,clip] {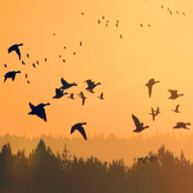}&
\includegraphics[width=0.12\linewidth,clip] 
{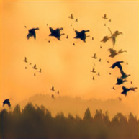}&
\includegraphics[width=0.12\linewidth,clip] {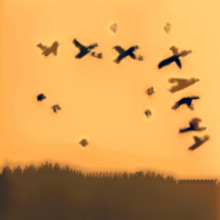}&
\includegraphics[width=0.12\linewidth,clip] {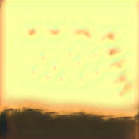}&
\includegraphics[width=0.12\linewidth,clip] {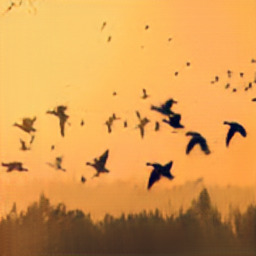}& 
\includegraphics[width=0.12\linewidth,clip] {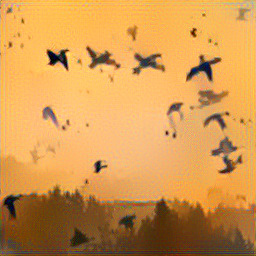}\\

\centering{\begin{small}\begin{turn}{90}\;\;\;\;\;\; \hspace{-5mm} \end{turn}\end{small}}&
\includegraphics[width=0.12\linewidth, clip]
{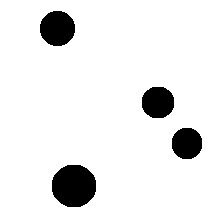}&
\includegraphics[width=0.12\linewidth,clip] {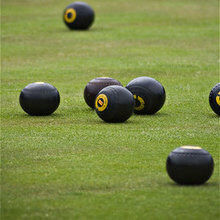}&
\includegraphics[width=0.12\linewidth,clip] 
{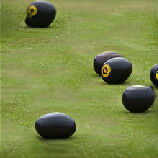}&
\includegraphics[width=0.12\linewidth,clip] {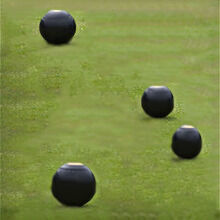}&
\includegraphics[width=0.12\linewidth,clip] {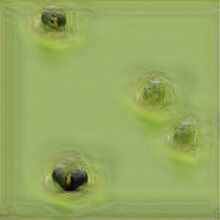}&
\includegraphics[width=0.12\linewidth,clip] {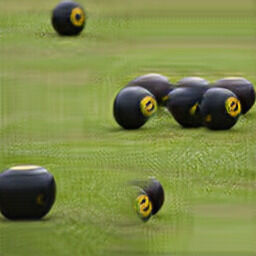}& 
\includegraphics[width=0.12\linewidth,clip] {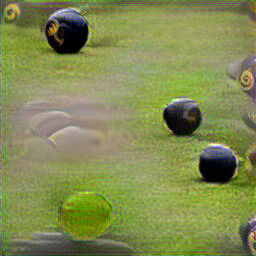}\\

\centering{\begin{small}\begin{turn}{90}\;\;\;\;\;\; \hspace{-5mm} \end{turn}\end{small}}&
\includegraphics[width=0.12\linewidth, clip]
{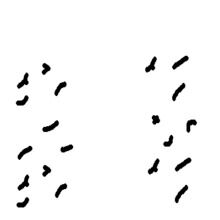}&
\includegraphics[width=0.12\linewidth,clip] {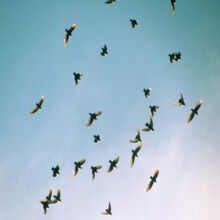}&
\includegraphics[width=0.12\linewidth,clip] 
{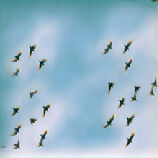}&
\includegraphics[width=0.12\linewidth,clip] {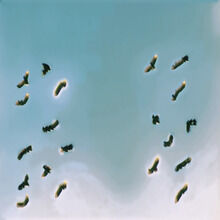}&
\includegraphics[width=0.12\linewidth,clip] {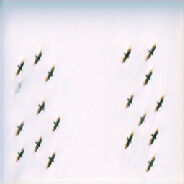}&
\includegraphics[width=0.12\linewidth,clip] {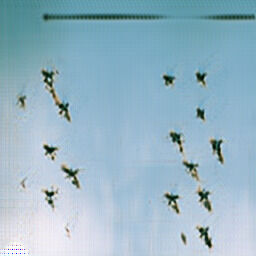}& 
\includegraphics[width=0.12\linewidth,clip] {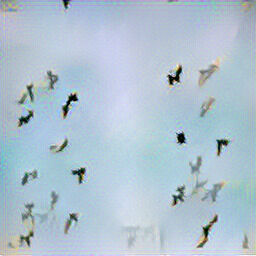}\\

\centering{\begin{small}\begin{turn}{90}\;\;\;\;\;\; \hspace{-5mm} \end{turn}\end{small}}&
\includegraphics[width=0.12\linewidth, clip]
{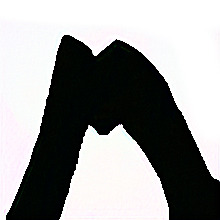}&
\includegraphics[width=0.12\linewidth,clip] {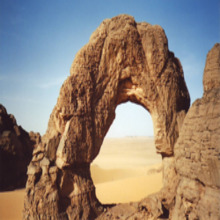}&
\includegraphics[width=0.12\linewidth,clip] 
{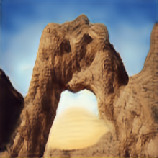}&
\includegraphics[width=0.12\linewidth,clip] {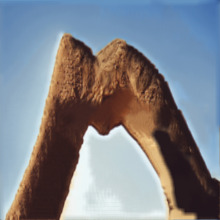}&
\includegraphics[width=0.12\linewidth,clip] {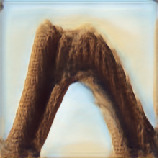}&
\includegraphics[width=0.12\linewidth,clip] {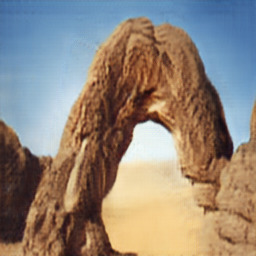}& 
\includegraphics[width=0.12\linewidth,clip] {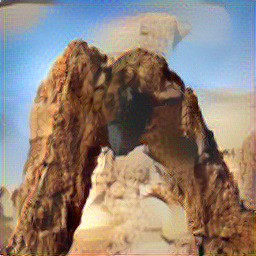}\\

\end{tabular}
\caption{From sketch  $A$, and image $B$, we generate a structurally aligned image.
\label{fig:sketchtoimage}}
\end{figure*}

\begin{figure*}
\centering
\includegraphics[width=\linewidth]{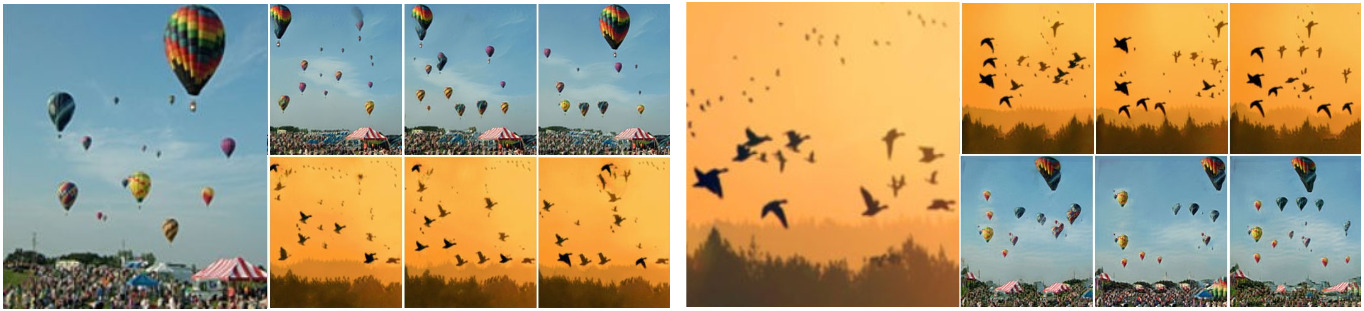} \\
(a) ~~~~~~~~~~~~~~~~~~~~~~~~~~~~~~~~~~~~~~~~~~~~~~~~~ (b)
\caption{(a) Left: Input image $A$ (hot air balloons). Right: Randomly generated samples $\overline{a}$ (top) and their translation $\overline{ab}$ (bottom). (b) As in (a) but for image $B$ (birds).}
\label{fig:random_samples}
\end{figure*}

\begin{figure*}
\setlength{\belowcaptionskip}{-5pt}
  \centering
  \begin{tabular}{c@{~}c:c@{~}c@{~}c@{~}c@{~}c}
&  Input & \textbf{Ours} & DIA ~\cite{deepimageanalogy} & SinGAN ~\cite{singan} & Cycle$^*$ ~\cite{CycleGAN2017} & Style ~\cite{Gatys_2016_CVPR} \\

\centering{\begin{small}\begin{turn}{90}\;\;\;\;\;\; \hspace{-28mm} Flower\&Bamboos to Starfish\&Seahorses \end{turn}\end{small}}&
\includegraphics[width=0.140\linewidth, clip]
{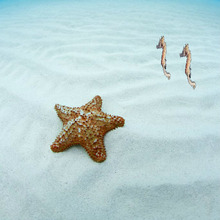}&
\includegraphics[width=0.140\linewidth,clip] {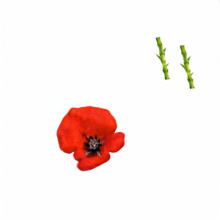}&
\includegraphics[width=0.140\linewidth,clip] 
{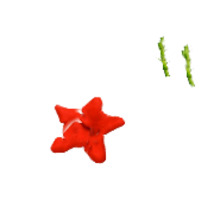}&
\includegraphics[width=0.140\linewidth,clip] {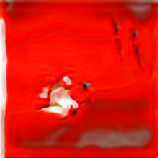}&
\includegraphics[width=0.140\linewidth,clip] {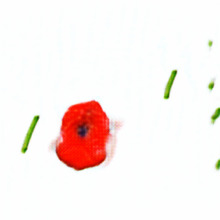}& 
\includegraphics[width=0.140\linewidth,clip] {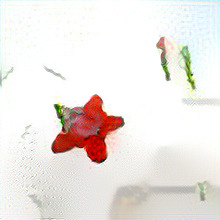}\\ 

&
\includegraphics[width=0.140\linewidth, clip]
{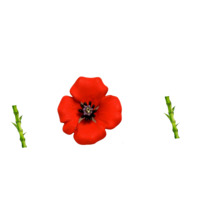}&
\includegraphics[width=0.140\linewidth,clip] {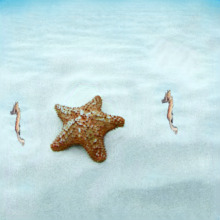}&
\includegraphics[width=0.140\linewidth,clip] 
{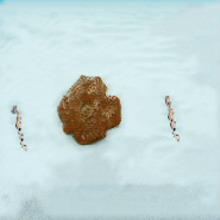}&
\includegraphics[width=0.140\linewidth,clip] {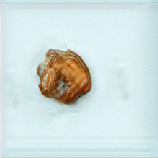}&
\includegraphics[width=0.140\linewidth,clip] {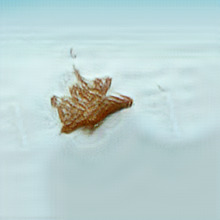}& 
\includegraphics[width=0.140\linewidth,clip] {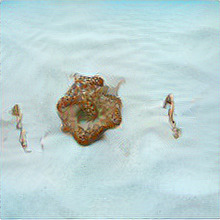}\\ 

\centering{\begin{small}\begin{turn}{90}\;\;\;\;\;\; \hspace{-30mm} Mountains\&Birds to Pyramids\&Balloons \end{turn}\end{small}}&
\includegraphics[width=0.140\linewidth, clip]
{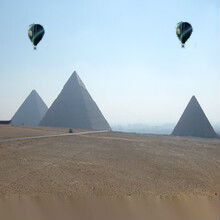}&
\includegraphics[width=0.140\linewidth,clip] {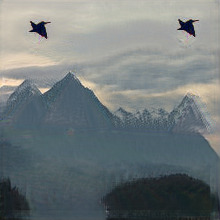}&
\includegraphics[width=0.140\linewidth,clip] 
{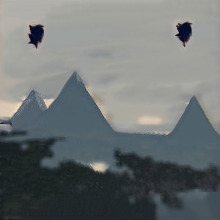}&
\includegraphics[width=0.140\linewidth,clip] {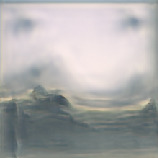}&
\includegraphics[width=0.140\linewidth,clip] {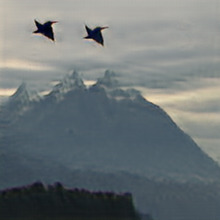}& 
\includegraphics[width=0.140\linewidth,clip] {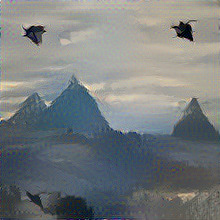}\\ 

&
\includegraphics[width=0.140\linewidth, clip]
{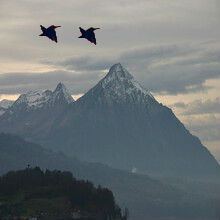}&
\includegraphics[width=0.140\linewidth,clip] {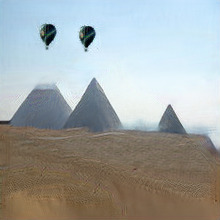}&
\includegraphics[width=0.140\linewidth,clip] 
{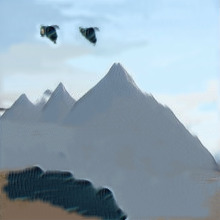}&
\includegraphics[width=0.140\linewidth,clip] {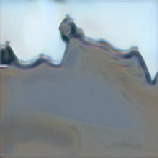}&
\includegraphics[width=0.140\linewidth,clip] {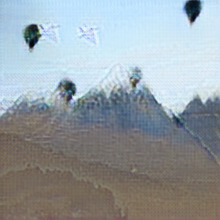}& 
\includegraphics[width=0.140\linewidth,clip] {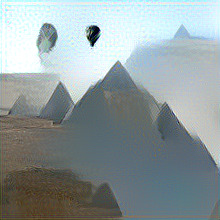}\\ 

\centering{\begin{small}\begin{turn}{90}\;\;\;\;\;\; \hspace{-24mm} Waves\&Wheel to Sky\&Ball \end{turn}\end{small}}&
\includegraphics[width=0.140\linewidth, clip]
{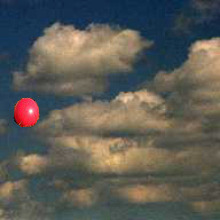}&
\includegraphics[width=0.140\linewidth,clip] {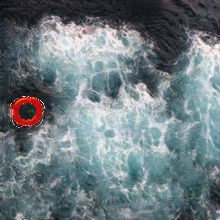}&
\includegraphics[width=0.140\linewidth,clip] 
{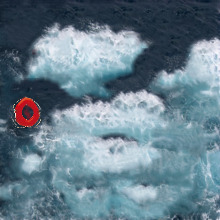}&
\includegraphics[width=0.140\linewidth,clip] {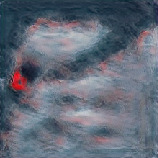}&
\includegraphics[width=0.140\linewidth,clip] {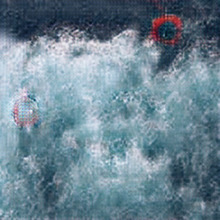}& 
\includegraphics[width=0.140\linewidth,clip] {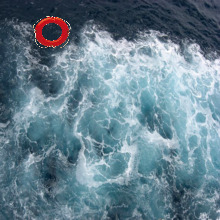}\\ 

&
\includegraphics[width=0.140\linewidth, clip]
{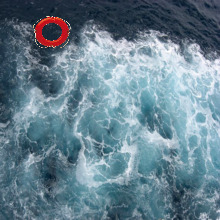}&
\includegraphics[width=0.140\linewidth,clip] {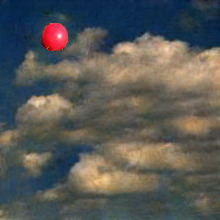}&
\includegraphics[width=0.140\linewidth,clip] 
{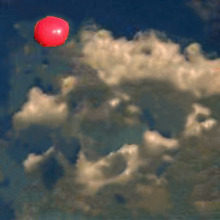}&
\includegraphics[width=0.140\linewidth,clip] {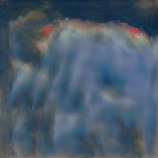}&
\includegraphics[width=0.140\linewidth,clip] {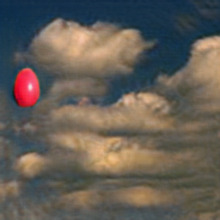}& 
\includegraphics[width=0.140\linewidth,clip] {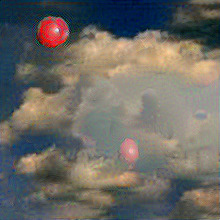}\\ 

\end{tabular}
\caption{Structural analogy for images contains two object classes.  
}
\label{fig:multi_result}
\end{figure*}

\paragraph*{Guided image synthesis}
We consider the mapping of a sketch drawn in black and white, $A$,  to a natural image, $B$, in Fig.~\ref{fig:sketchtoimage}. SinGAN~\cite{singan} performs a similar application, but requires $A$ to be of a similar patch distribution at a low scale of $B$. We do not make such assumption. 
For evaluation, we conduct a user study for realism (C1) and alignment (C2) as in Sec.~\ref{sec:quantitative}. 
As shown qualitatively in Fig.~\ref{fig:sketchtoimage},  DIA~\cite{deepimageanalogy} does not change the image structure, hence it gets a perfect alignment score, but its images are not realistic. CycleGAN has similar realism score to ours, but exhibits mode collapse and so its alignment score is significantly lower. 

\begin{figure*}
\smallskip
  \centering
  \begin{tabular}{c@{~}c@{~}c:c@{~}c@{~}c@{~}c}
&  Style ($B$) & Content ($A$) & \textbf{Ours} & DIA ~\cite{deepimageanalogy} & AdaIn ~\cite{adain} & Style ~\cite{Gatys_2016_CVPR} \\

\centering{\begin{small}\begin{turn}{90}\;\;\;\;\;\;
\end{turn}\end{small}}&
\includegraphics[width=0.12\linewidth, clip]
{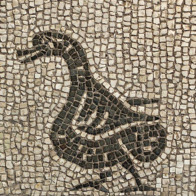}&
\includegraphics[width=0.12\linewidth,clip] {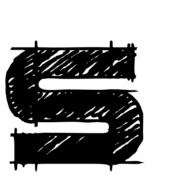}&
\includegraphics[width=0.12\linewidth,clip] 
{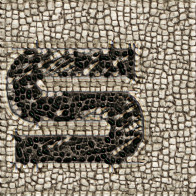}&
\includegraphics[width=0.12\linewidth,clip] {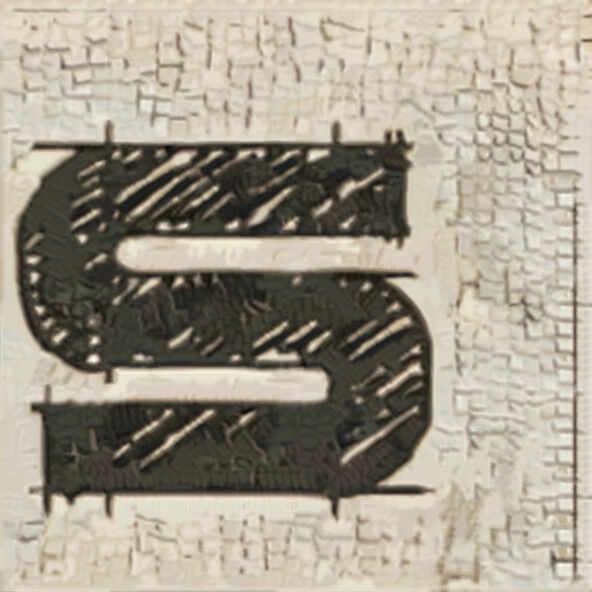}&
\includegraphics[width=0.12\linewidth,clip] {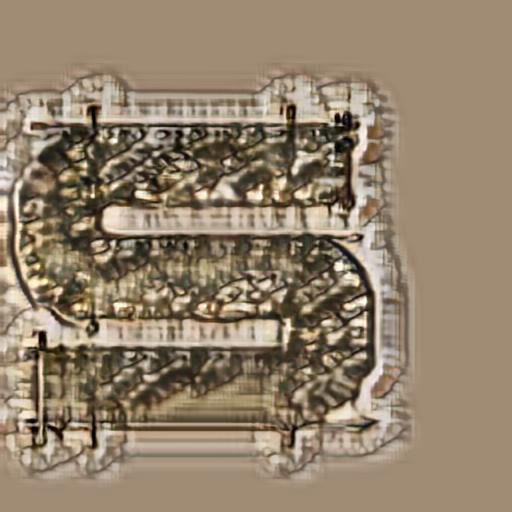}& 
\includegraphics[width=0.12\linewidth,clip] {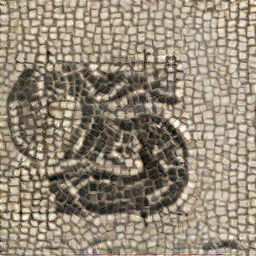}\\

&
\includegraphics[width=0.12\linewidth,clip] {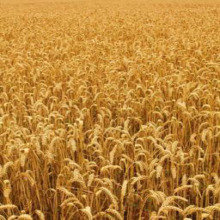}&
\includegraphics[width=0.12\linewidth,clip] {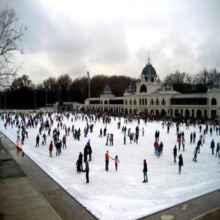}&
\includegraphics[width=0.12\linewidth,clip] 
{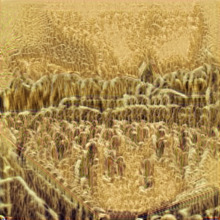}&
\includegraphics[width=0.12\linewidth,clip] {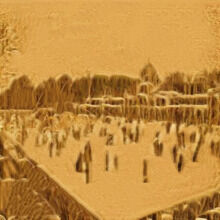}&
\includegraphics[width=0.12\linewidth,clip] {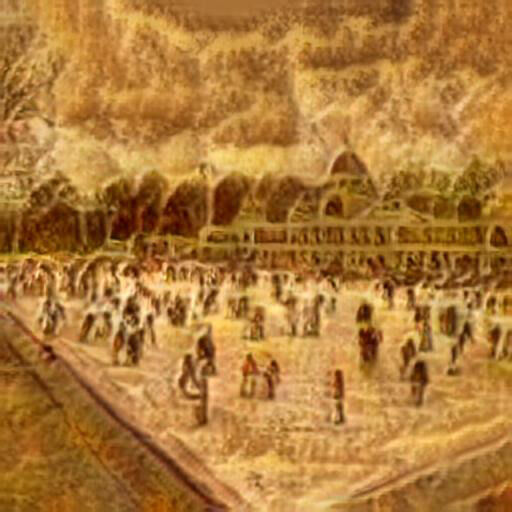}& 
\includegraphics[width=0.12\linewidth,clip] {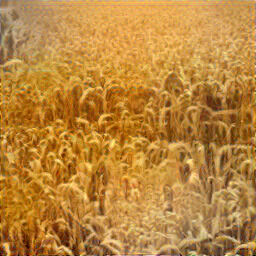}\\ 

& Texture ($B$) & Content ($A$)&  \textbf{Ours} & DIA ~\cite{deepimageanalogy} & AdaIn ~\cite{adain} & Style ~\cite{Gatys_2016_CVPR} \\

\centering{\begin{small}\begin{turn}{90}\;\;\;\;\;\; \end{turn}\end{small}}&
\includegraphics[width=0.12\linewidth,clip] {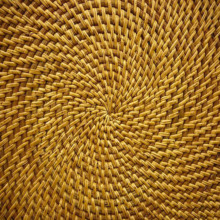}&
\includegraphics[width=0.12\linewidth,clip] {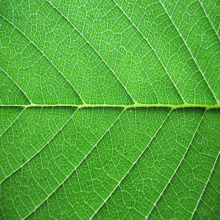}&
\includegraphics[width=0.12\linewidth,clip] 
{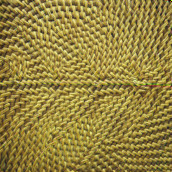}&
\includegraphics[width=0.12\linewidth,clip] {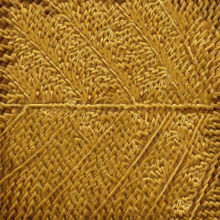}&
\includegraphics[width=0.12\linewidth,clip] {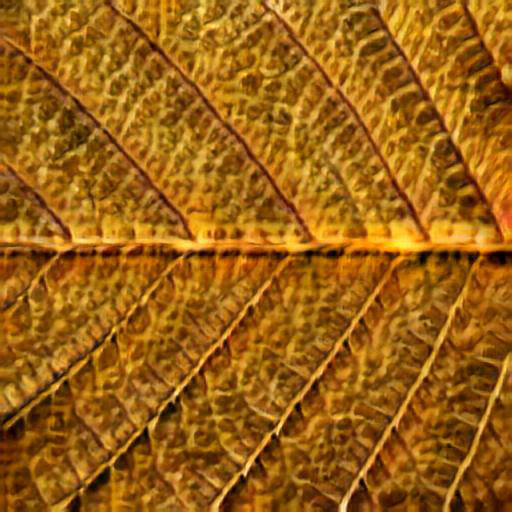}&
\includegraphics[width=0.12\linewidth,clip] 
{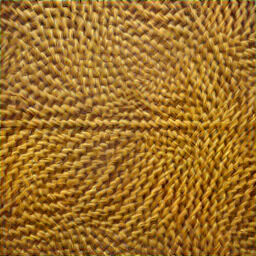}\\

\centering{\begin{small}\begin{turn}{90}\;\;\;\;\;\; \end{turn}\end{small}}&
\includegraphics[width=0.12\linewidth, clip]
{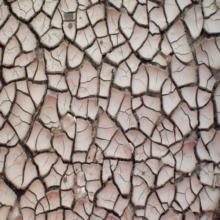}&
\includegraphics[width=0.12\linewidth,clip] {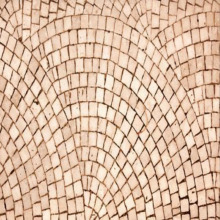}&
\includegraphics[width=0.12\linewidth,clip] 
{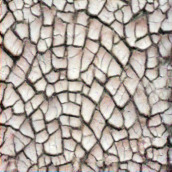}&
\includegraphics[width=0.12\linewidth,clip] {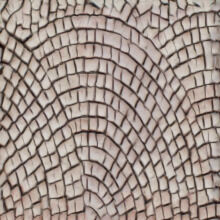}&
\includegraphics[width=0.12\linewidth,clip] {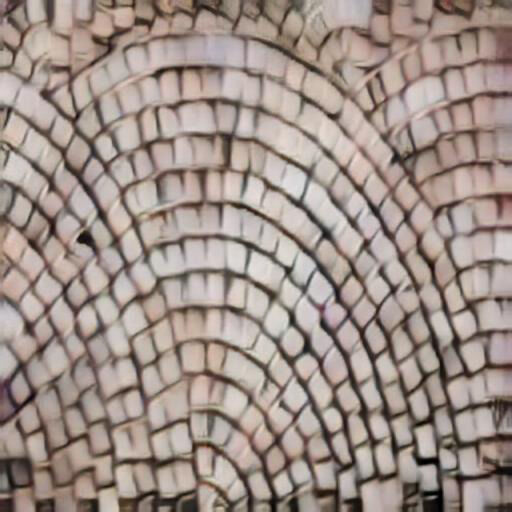}&
\includegraphics[width=0.12\linewidth,clip] 
{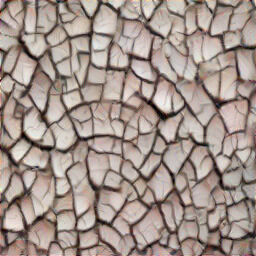}\\ 
\end{tabular}
\caption{An illustration of our method for the task of style and texture transfer.\label{fig:style}}

\end{figure*}

\paragraph*{Style and texture Transfer}
Our method can also adapt the style or texture of a source image to a target one. This is illustrated compared against baseline methods in Fig.~\ref{fig:style}. 

\begin{figure*}
\centering
  \begin{tabular}{c@{~}c@{~}c:@{~}c@{~}c@{~}c@{~}c@{~}c@{~}c}
  & \tiny{Style ($B$)} & \tiny{Content ($A$)} & \tiny{\textbf{Ours}} & 
  \tiny{DIA~\cite{deepimageanalogy}}  & \tiny{SinGAN~\cite{singan}} & \tiny{Cycle$^*$~\cite{CycleGAN2017}} &  \tiny{Style~\cite{Gatys_2016_CVPR}} & \tiny{SM-GAN\cite{ font1}}  \\

\centering{\begin{small}\begin{turn}{90}\;\;\;\;\;\; \end{turn}\end{small}}&
\includegraphics[width=0.11\linewidth, clip]
{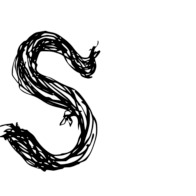}&
\includegraphics[width=0.11\linewidth,clip] {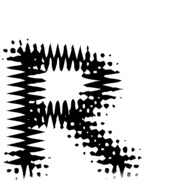}&
\includegraphics[width=0.11\linewidth,clip] 
{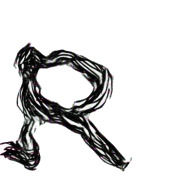}&
\includegraphics[width=0.11\linewidth,clip] {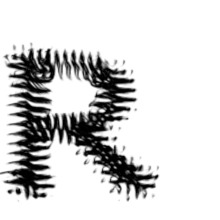}& \includegraphics[width=0.11\linewidth,clip] {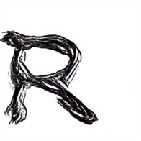}&
\includegraphics[width=0.11\linewidth,clip] {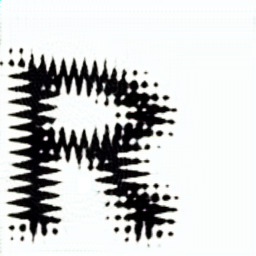}&
\includegraphics[width=0.11\linewidth,clip] {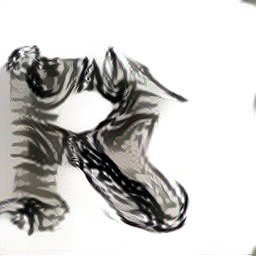}&
\includegraphics[width=0.11\linewidth,clip] {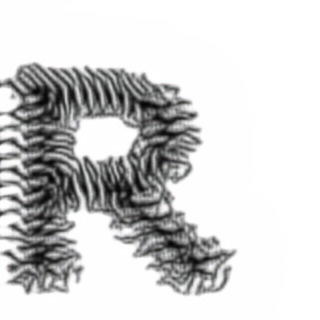}\\

\centering{\begin{small}\begin{turn}{90}\;\;\;\;\;\; \end{turn}\end{small}}&
\includegraphics[width=0.11\linewidth, clip]
{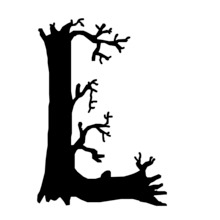}&
\includegraphics[width=0.11\linewidth,clip] {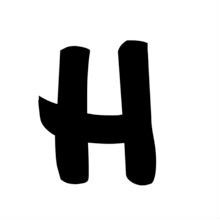}&
\includegraphics[width=0.11\linewidth,clip] 
{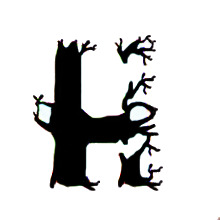}&
\includegraphics[width=0.11\linewidth,clip] {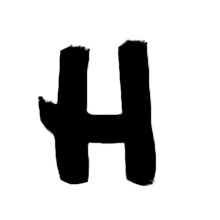}& \includegraphics[width=0.11\linewidth,clip] {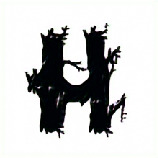}&
\includegraphics[width=0.11\linewidth,clip] {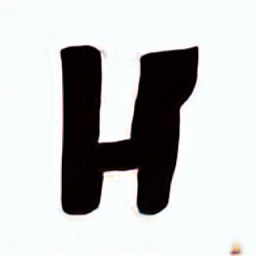}&
\includegraphics[width=0.11\linewidth,clip] {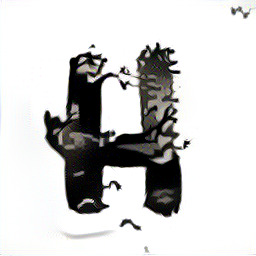}&
\includegraphics[width=0.11\linewidth,clip] {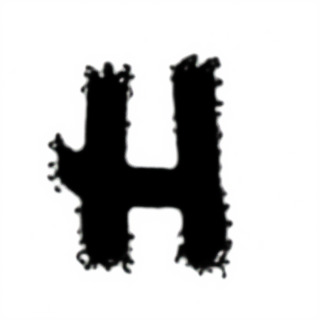}\\

\centering{\begin{small}\begin{turn}{90}\;\;\;\;\;\; \end{turn}\end{small}}&
\includegraphics[width=0.11\linewidth, clip]
{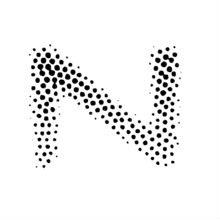}&
\includegraphics[width=0.11\linewidth,clip] {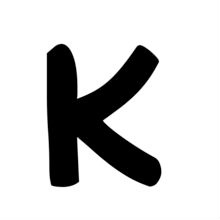}&
\includegraphics[width=0.11\linewidth,clip] 
{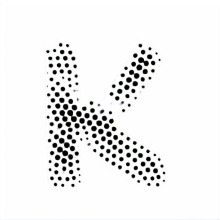}&
\includegraphics[width=0.11\linewidth,clip] {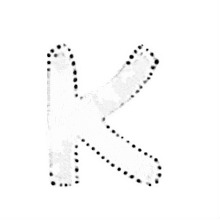}& \includegraphics[width=0.11\linewidth,clip] {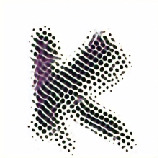}&
\includegraphics[width=0.11\linewidth,clip] {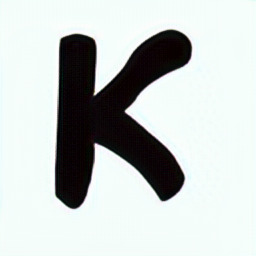}&
\includegraphics[width=0.11\linewidth,clip] {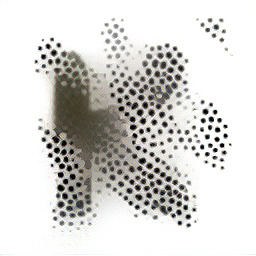}&
\includegraphics[width=0.11\linewidth,clip] {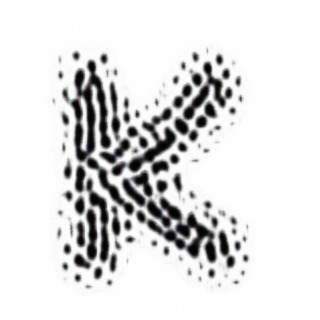}\\
\end{tabular}
\caption{Text translation.  
We also compare to the text transfer method of SM-GAN~\cite{font1}.\label{fig:texttranslation}}
\end{figure*}

\begin{figure*}
\centering
\includegraphics[width=0.7\linewidth]{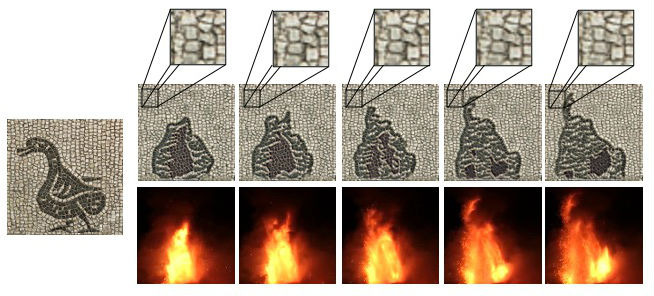} \\
~~~~~~~~~~~~~~~~~(a) ~~~~~~~~~~~~~~~~~~~~~~~~~~~~~~~~~~~~~~~~~~~~~~~~~~~~~~~~~~~~~~~~~~~~~~ (b) ~~~~~~~~~~~~~~~~~~~~~~~~~~~~~~~~~~~~~~~~~~~~~~~~~~~~~~~~~~~~~~~~~~ 
\caption{Video translation. We translate between a source video of a volcano and mosaic target image. (a) Target image (b) Bottom row shows the input frames $f_i$. Middle row shows their translation $v_i$. Upper row shows a zoomed in part of the frame, where identical for all translated frames since there is no motion in this part. Frames $65, 85, 105, 135, 155$ are chosen for this illustration.}
\label{fig:videotranslation}
\end{figure*}

\begin{figure*}
  \centering
  \centering
  \begin{tabular}{c@{~}c:c@{~}c@{~}c@{~}c@{~}c@{~}c@{~}c@{~}c@{~}c@{~}c}
  & Input & (a) & (b) & (c) & (d) & (e) & (f) & (g) & (h) & (i) \\
  
\centering{\begin{small}\begin{turn}{90}\;\;\;\;\;\; \end{turn}\end{small}}&
\includegraphics[width=0.09\linewidth, clip]
{image2image_results/pumpkin_real_b.jpg}&
\includegraphics[width=0.09\linewidth,clip] 
{image2image_results/pumpkin_b2a_ours.jpg}&

\includegraphics[width=0.09\linewidth,clip] 
{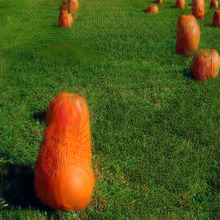}&
\includegraphics[width=0.09\linewidth,clip] 
{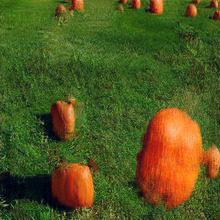}&
\includegraphics[width=0.09\linewidth,clip] 
{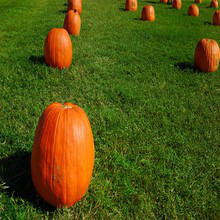}&
\includegraphics[width=0.09\linewidth,clip] 
{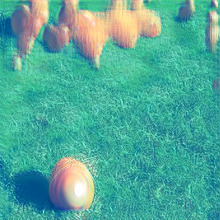}&
\includegraphics[width=0.09\linewidth,clip] 
{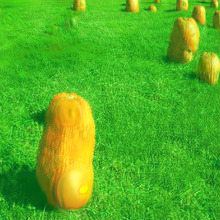}&
\includegraphics[width=0.09\linewidth,clip] 
{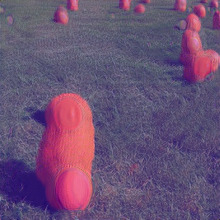}&
\includegraphics[width=0.09\linewidth,clip] 
{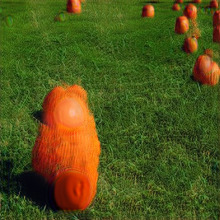}&
\includegraphics[width=0.09\linewidth,clip] 
{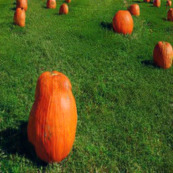}\\

\centering{\begin{small}\begin{turn}{90}\;\;\;\;\;\; \end{turn}\end{small}}&
\includegraphics[width=0.09\linewidth,clip] {image2image_results/pumpkin_real_a.jpg}&
\includegraphics[width=0.09\linewidth,clip] 
{image2image_results/pumpkin_a2b_ours.jpg}&

\includegraphics[width=0.09\linewidth,clip] 
{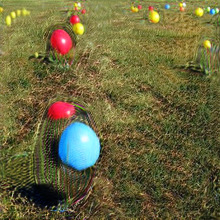}&
\includegraphics[width=0.09\linewidth,clip] 
{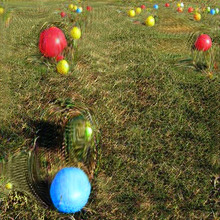}&
\includegraphics[width=0.09\linewidth,clip] 
{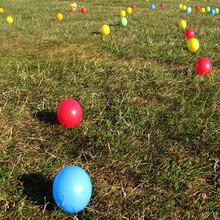}&
\includegraphics[width=0.09\linewidth,clip] 
{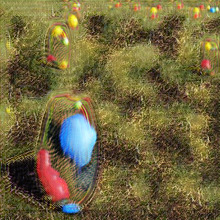}&
\includegraphics[width=0.09\linewidth,clip] 
{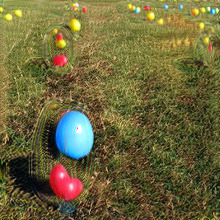}&
\includegraphics[width=0.09\linewidth,clip] 
{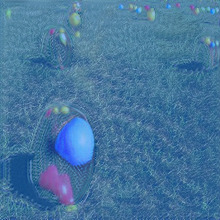}&
\includegraphics[width=0.09\linewidth,clip] 
{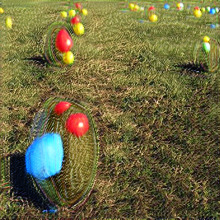}&
\includegraphics[width=0.09\linewidth,clip] 
{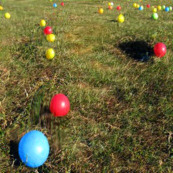}\\

\end{tabular}
\caption{An ablation analysis for the translation of balls to pumpkins. (a) original translation. In (b), the cycle loss (Eq.~\ref{eq:cycle_loss}) is omitted for all but scale $K-1$. (c), no cycle loss is employed at all. (d), $G^n_B$ in Eq.\ref{eq:cond1_res}  is changed to accept $\overline{a}_n$ as well as $\uparrow\overline{ab}_{n-1}$ --- the translation from $A$ to $B$ at scale $n-1$. In (e), no reconstruction loss (Eq.~\ref{eq:recon_loss}) is used. In (f) a residual generator (see Sec.~\ref{sec:arch}) is used for all scales while in (g) no residual training is used at all. In (h), we do not initialize the networks of scale $n$ with those of scale $n-1$ (see Sec.~\ref{sec:arch}). In (i), separate networks (no shared weights) are used for conditional and unconditional generation. See Sec.~\ref{sec:ablation}  for additional discussion.}
\label{fig:ablation}
\end{figure*}

\paragraph*{Text translation}
In the task of text translation, we are given two images of text, and are asked to transfer the style characteristics of the first image to the second one. This is not a traditional style transfer task, since the style is embedded in the geometry of the image, rather than its appearance. Fig.~\ref{fig:texttranslation} illustrates these mappings, and indeed our method correctly transfers the style characteristics of the input text including fine details, while DIA and CycleGAN fail in all cases. Style transfer of Gatys et al.~\cite{Gatys_2016_CVPR} and SM-GAN~\cite{font1} only partially transfer the style characteristics. SinGAN~\cite{singan} performs most closely to us, but the generated texture does not resemble the input texture as well.

\paragraph*{Video translation}

In the task of video translation, we are given an input source video and a target image. We are asked to generate a video, in which every frame is structurally aligned to the corresponding frame in the source video, but belongs to the patch distribution of the target image. 
Consider a source video consisting of frames $f_1, \dots, f_k$. We would like to incorporate these frames in training, to further improve the translation quality. Therefore, in each iteration, $A$ is randomly chosen to be one of the frames, $f_i$. $B$ is chosen to be the target image. 
At inference time, we translate one frame at a time to construct an analogous video. For frame $f_i$ we first generate a refined frame $f_i^r$ as follows:
\begin{align}
f_i^r = G^N_A(z_N+\uparrow G^{N-1}_A(z_{N-1} + f_{i,{N-1}}))
\end{align}
Where $f_{i,{N-1}}$ is frame $f_i$ resized to scale $N-1$. The translated frame $v_i$ is given by $v_i = G^N_B(f_i^r)$.

We find that using $f_i^r$ results in a superior quality then using $f_i$. In addition, $z_n$ and $z_{n-1}$ are fixed Gaussian noise which is used for all frames $f_i$. We find that using a fixed noise helps generate temporally consistent frames. For example, as shown in Fig.~\ref{fig:videotranslation}, some part of the frame remains fixed, while the rest changes in a manner which is consistent with the movement of the source volcano video. Furthermore, in some cases, we found that quantization of the colors of the frames prior to the translation improves the quality of the result. 
Full videos are provided in the project webpage: \url{https://sagiebenaim.github.io/structural-analogy/}.

\subsection{Baselines}
\label{sec:baselines}

Baseline methods were trained in their original configuration using official public code. For SinGAN baseline used in Sec.~4.1, we insert a downsampled version of image $A$ at a coarse scale of $B$’s pretrained model. This coarse scale is chosen to be $2$. We found that using a value of $1$ resulted in no local structure being preserved, while for a scale larger then $2$, generation quality was worse. For the CycleGAN model (CycleGAN trained on images generated using two SinGANs), we use the original configuration of SinGAN for both $A$ and $B$ (such as minimum image size of $25px$ and maximum image size of $250px$) and generated random images by applying random noise at all scales, to generated maximum variability. 
In AdaIn~\cite{adain}, a model is trained with many style and content images. As the original implementation uses only art images for style, we added images which are not artistic, as used in our method.

\subsection{Ablation Analysis}
\label{sec:ablation}
An ablation analysis is presented in Fig.~\ref{fig:ablation} for the case of translating from pumpkins ($A$) to balls ($B)$: (a) is the original translation as in Fig~\ref{fig:images_objects}. In (b), the cycle loss (Eq.~\ref{eq:cycle_loss}) is omitted for all but scale $K-1$ and in (c), no cycle loss is employed at all, both resulting in less alignment. In (b), for example, the two balls are mapped to a single large pumpkin, while in (c) the generated pumpkin on the bottom right is unaligned.  In (d), $G^n_B$ in Eq.\ref{eq:cond1_res}  is changed to accept $\overline{a}_n$ as well as $\uparrow\overline{ab}_{n-1}$ --- the translation from $A$ to $B$ at scale $n-1$. This results in mode collapse, in which the source image $A$ is generated. 
In (e), no reconstruction loss (Eq.~\ref{eq:recon_loss}) is used, which results in no alignment and worse generation quality. Generated images also look tinted. In (f) a residual generator (see Sec.~\ref{sec:arch}) is used for all scales while in (g) no residual training is used at all. In both cases alignment is reduced and images look tinted. In (h), we do not initialize the networks of scale $n$ with those of scale $n-1$ (see Sec.~\ref{sec:arch}), which results in a worse alignment and unrealistic images. In (i), two separate networks (no shared weights) are used for the conditional and unconditional generation at every scale. That is the generation of $\overline{a}_n$ is with a network $G^n_A$ while that of $\overline{ba}_n$ and $\overline{aba}_n$ is with a network $G'^n_A$ that has the same architecture but is learned separately to $G^n_A$. As can be seen, alignment is impaired.  

In the appendix Fig.~\ref{fig:k_abl2} we present the result of using different $K$ values for conditional generation. 
Using a low value of $K$ results in unaligned solutions. For example, the orange ball is misplaced and the resulting image is not aligned to the input one. For higher values of $K$ the orange ball is correctly placed. 
Using residual training for most layers ($K=12$) results in the correct alignment. In the presented example, three balls in the resulting image appear against three balls in the input.
Using residual training for all networks (i.e $K=13$), the method is unable to produce the correct texture.  
In the appendix Fig.~\ref{fig:k_abl3} we  present the result of using different $K$ values for an unconditional generation; Specifically, the reconstruction of the input in the last scale $N$. As can be seen, reconstruction of the input is almost perfect, other than the $K=13$, where the texture differs from the input image.

\subsection{Inference Ablation}
\label{sec:inference_ablation}

Sec.~\ref{sec:inference} describes our inference procedure, where $S$ is an hyperparameter. The effect of choosing a different value for $S$ is shown in the appendix Fig.~\ref{fig:inj}. As can be seen, the best result is using $S=7$. For other $S$ values the image is blurrier (e.g. $S<4$), contains artifacts (e.g. $S=4,5$) or less aligned (e.g. $S=6$), i.e. different structure is present.

As an alternative inference procedure, we consider performing the translation between domain $A$ to domain $B$ before the final scale $N$. We first inject image $A$ at scale $S$, getting $a_S$. We then upsample it to a higher scale $S' < N$ resulting in sample $a^*_{S'}$. Next, we  map $a^*_{S'}$ to the target domain, resulting in $\overline{ab}_{S'}$. Finally we upsample $\overline{ab}_{S'}$ to scale $N$. As shown in in the appendix Fig.~\ref{fig:switch}, there is no advantage of performing the mapping before the last scale $N$. 

\subsection{Computational Time}

As a further ablation, we consider the number of iterations and clock-time time required to train on a given pair of images. For all our examples, we used $10000$ iterations for each scale, resulting in a total clock-time of $8.5$ hours on a single TITAN X Nvidia GPU.
Using less than $10000$ iterations results in a degradation of quality, as can be seen in the appendix Fig.~\ref{fig:computational_time}.

\subsection{Failure Cases}
\label{sec:failure_cases}
The input to our method is only a single pair of images, without any additional guidance, such as perceptual loss or additional images. Therefore, it lacks a semantic understanding of the world. For example, given an image of a cat with two eyes, it is impossible to deduce that every tiger has exactly two eyes since we do not have further examples to support this hypothesis. Under this limitation, it is clear that our method does not perform well when the target domain requires semantic manipulation. Hence, as illustrated in the appendix Fig.~\ref{fig:fail}, when translating between a cat and a tiger, the eyes are not preserved. See the appendix Fig.~\ref{fig:fail} for additional failure cases.

\section{Conclusions}

Up until recently, the problem of unsupervised image-to-image translation, without additional images from the same domains, was not considered possible. 
Our method is the first to consider only a single pair of images and to successfully generate {structural analogies}, even if the structure, or semantics, of the objects in the images differ significantly (e.g., bird and balloons). 
By considering {structural analogies}, our method goes beyond style transfer, and can change the structure of local objects or parts by matching the internal patch statistics of the source and target images.
Our method can also be applied in other conditional generation tasks such as guided image synthesis, style and texture transfer and even text and video translation. Going forward, we hope to use our method in other conditional generation tasks, which traditionally require costly supervision, for example, for one-shot semantic segmentation from a single image. 
\newline
\newline
\newline

\clearpage

\printbibliography

\appendix

\begin{figure*}
  \centering
  \begin{tabular}{c@{~}c:c@{~}c@{~}c@{~}c@{~}c}
&  Input & \textbf{Ours} & DIA~\cite{deepimageanalogy} & SinGAN~\cite{singan} & Cycle$^*$~\cite{CycleGAN2017} & Style~\cite{Gatys_2016_CVPR} \\

\centering{\begin{small}\begin{turn}{90}\;\;\;\;\;\; \hspace{-18mm} Steps2Steps \end{turn}\end{small}}&

\includegraphics[width=0.140\linewidth, clip]
{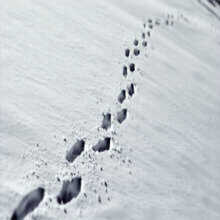}&
\includegraphics[width=0.140\linewidth,clip] {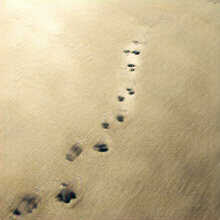}&
\includegraphics[width=0.140\linewidth,clip] 
{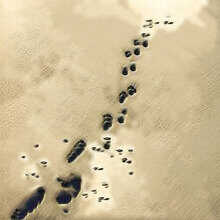}&
\includegraphics[width=0.140\linewidth,clip] {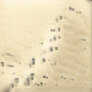}&
\includegraphics[width=0.140\linewidth,clip] {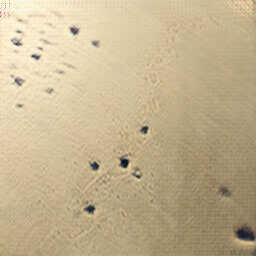}& 
\includegraphics[width=0.140\linewidth,clip] {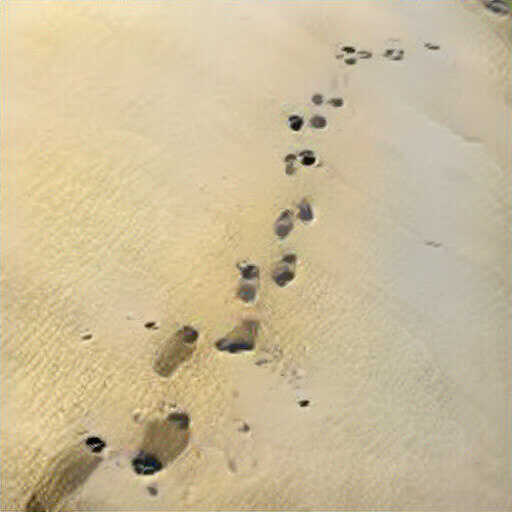}\\ 

&
\includegraphics[width=0.140\linewidth, clip]
{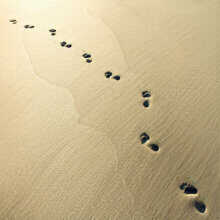}&
\includegraphics[width=0.140\linewidth,clip] {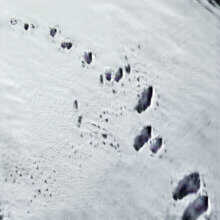}&
\includegraphics[width=0.140\linewidth,clip] 
{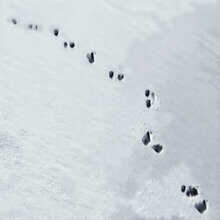}&
\includegraphics[width=0.140\linewidth,clip] {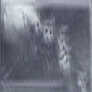}&
\includegraphics[width=0.140\linewidth,clip] {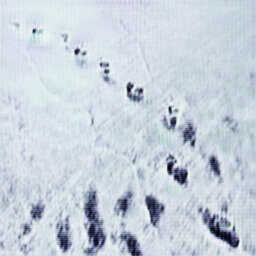}& 
\includegraphics[width=0.140\linewidth,clip] {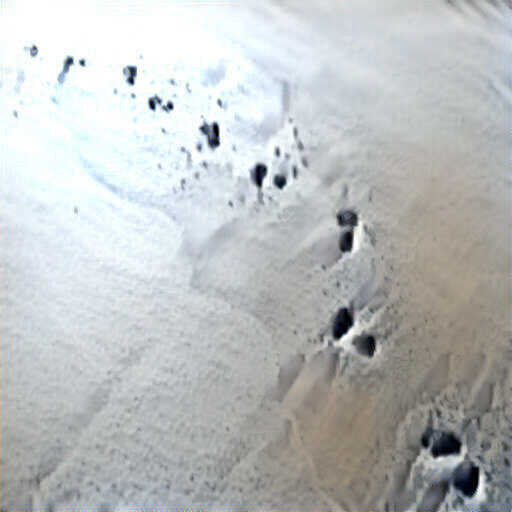}\\ 

\centering{\begin{small}\begin{turn}{90}\;\;\;\;\;\; \hspace{-18mm} Flowers2Feathers \end{turn}\end{small}}&
\includegraphics[width=0.140\linewidth, clip]
{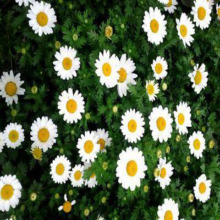}&
\includegraphics[width=0.140\linewidth,clip] {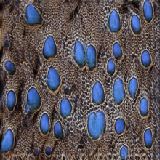}&
\includegraphics[width=0.140\linewidth,clip] 
{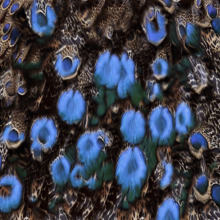}&
\includegraphics[width=0.140\linewidth,clip] {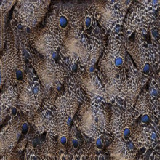}&
\includegraphics[width=0.140\linewidth,clip] {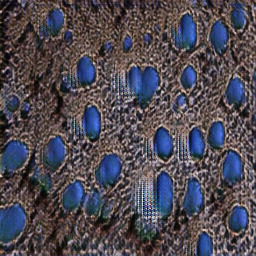}& 
\includegraphics[width=0.140\linewidth,clip] {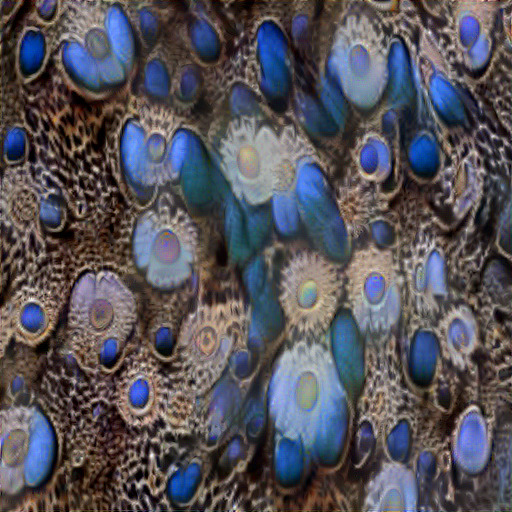} \\

&
\includegraphics[width=0.140\linewidth, clip]
{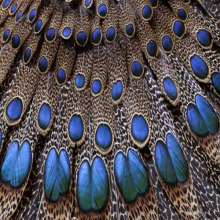}&
\includegraphics[width=0.140\linewidth,clip] {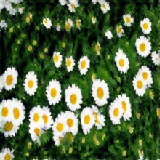}&
\includegraphics[width=0.140\linewidth,clip] 
{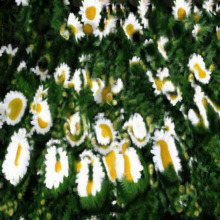}&
\includegraphics[width=0.140\linewidth,clip] {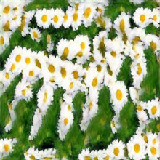}&
\includegraphics[width=0.140\linewidth,clip] {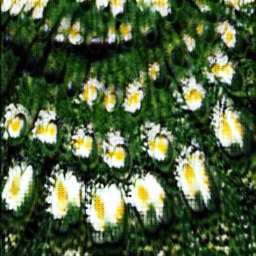}& 
\includegraphics[width=0.140\linewidth,clip] {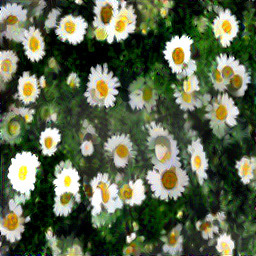}\\

\centering{\begin{small}\begin{turn}{90}\;\;\;\;\;\; \hspace{-18mm} Flowers2Cows \end{turn}\end{small}}&
\includegraphics[width=0.140\linewidth, clip]
{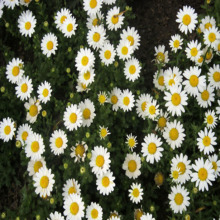}&
\includegraphics[width=0.140\linewidth,clip] {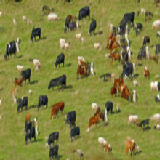}&
\includegraphics[width=0.140\linewidth,clip] 
{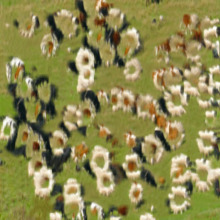}&
\includegraphics[width=0.140\linewidth,clip] {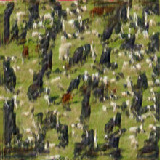}&
\includegraphics[width=0.140\linewidth,clip] {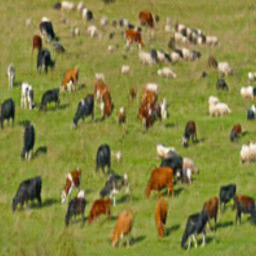}& 
\includegraphics[width=0.140\linewidth,clip] {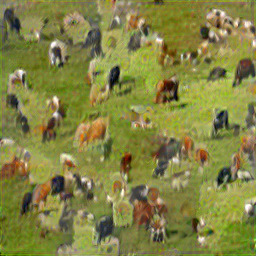} \\

&
\includegraphics[width=0.140\linewidth, clip]
{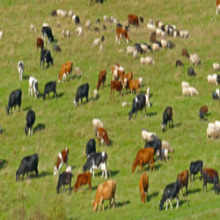}&
\includegraphics[width=0.140\linewidth,clip] {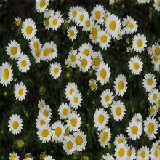}&
\includegraphics[width=0.140\linewidth,clip] 
{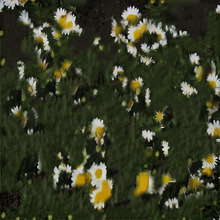}&
\includegraphics[width=0.140\linewidth,clip] {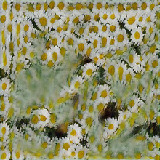}&
\includegraphics[width=0.140\linewidth,clip] {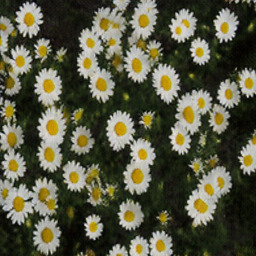}& 
\includegraphics[width=0.140\linewidth,clip] {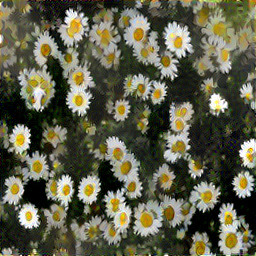}\\

\centering{\begin{small}\begin{turn}{90}\;\;\;\;\;\; \hspace{-16mm} Waves2Sky \end{turn}\end{small}}&
\includegraphics[width=0.140\linewidth, clip]
{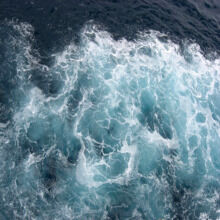}&
\includegraphics[width=0.140\linewidth,clip] {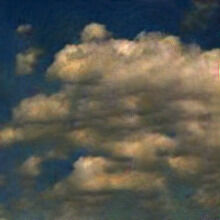}&
\includegraphics[width=0.140\linewidth,clip] 
{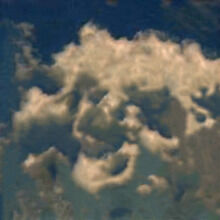}&
\includegraphics[width=0.140\linewidth,clip] {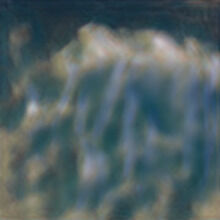}&
\includegraphics[width=0.140\linewidth,clip] {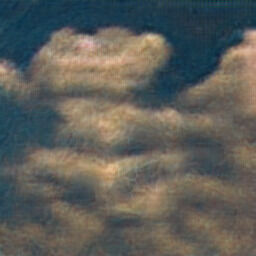}& 
\includegraphics[width=0.140\linewidth,clip] {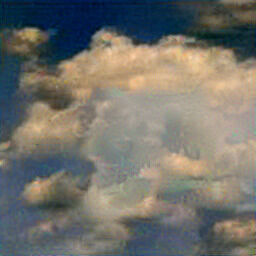}\\ 

&
\includegraphics[width=0.140\linewidth, clip]
{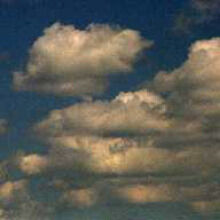}&
\includegraphics[width=0.140\linewidth,clip] {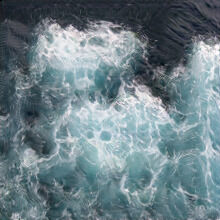}&
\includegraphics[width=0.140\linewidth,clip] 
{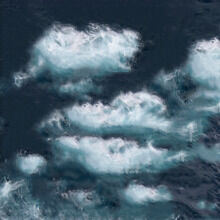}&
\includegraphics[width=0.140\linewidth,clip] {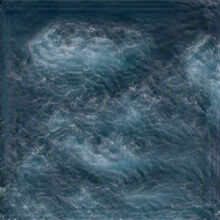}&
\includegraphics[width=0.140\linewidth,clip] {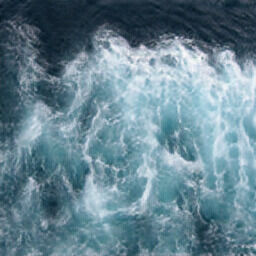}& 
\includegraphics[width=0.140\linewidth,clip] {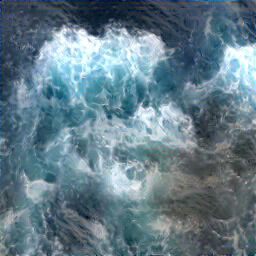}\\ 

\end{tabular}
\caption{Additional structural alignment results.  
}
\label{fig:additional2}
\end{figure*}

\begin{figure*}
  \centering
  \begin{tabular}{c@{~}c:c@{~}c@{~}c@{~}c@{~}c}
&  Input & \textbf{Ours} & DIA~\cite{deepimageanalogy} & SinGAN~\cite{singan} & Cycle$^*$~\cite{CycleGAN2017} & Style~\cite{Gatys_2016_CVPR} \\

\centering{\begin{small}\begin{turn}{90}\;\;\;\;\;\; \hspace{-18mm} Oranges2Apples \end{turn}\end{small}}&

\includegraphics[width=0.140\linewidth, clip]
{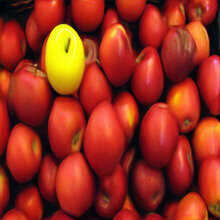}&
\includegraphics[width=0.140\linewidth,clip] {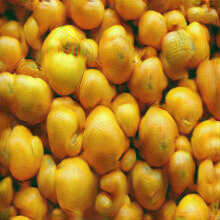}&
\includegraphics[width=0.140\linewidth,clip] 
{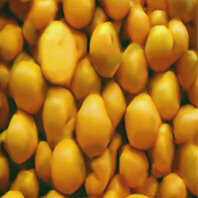}&
\includegraphics[width=0.140\linewidth,clip] {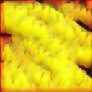}&
\includegraphics[width=0.140\linewidth,clip] {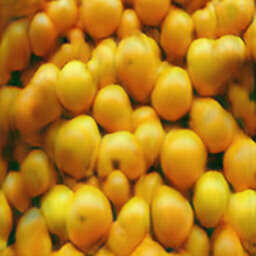}& 
\includegraphics[width=0.140\linewidth,clip] {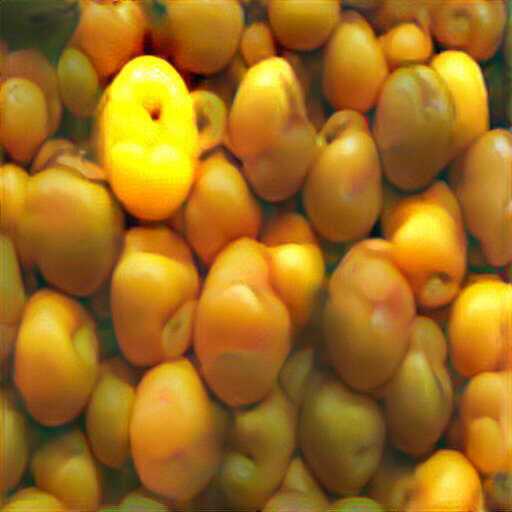}\\ 

&
\includegraphics[width=0.140\linewidth, clip]
{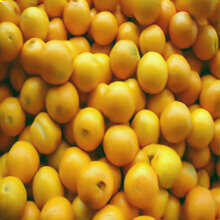}&
\includegraphics[width=0.140\linewidth,clip] {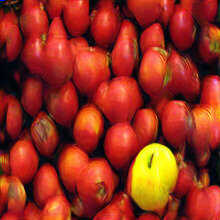}&
\includegraphics[width=0.140\linewidth,clip] 
{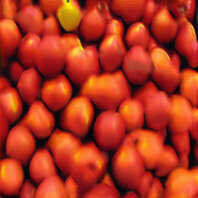}&
\includegraphics[width=0.140\linewidth,clip] {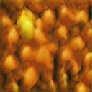}&
\includegraphics[width=0.140\linewidth,clip] {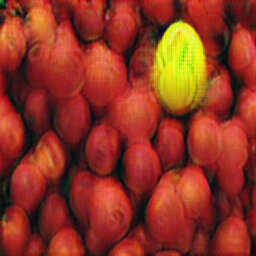}& 
\includegraphics[width=0.140\linewidth,clip] {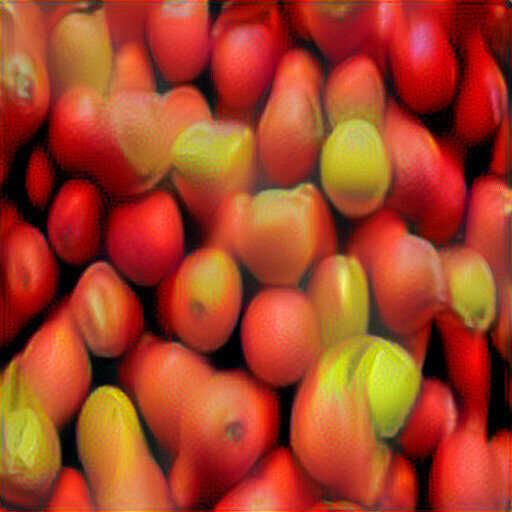}\\ 

\centering{\begin{small}\begin{turn}{90}\;\;\;\;\;\; \hspace{-18mm} Snow2Ice \end{turn}\end{small}}&

\includegraphics[width=0.140\linewidth, clip]
{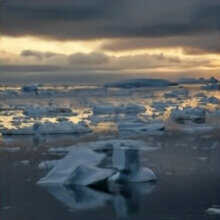}&
\includegraphics[width=0.140\linewidth,clip] {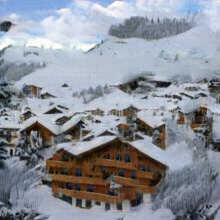}&
\includegraphics[width=0.140\linewidth,clip] 
{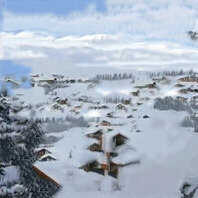}&
\includegraphics[width=0.140\linewidth,clip] {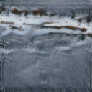}&
\includegraphics[width=0.140\linewidth,clip] {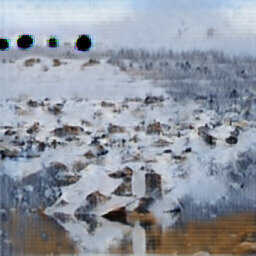}& 
\includegraphics[width=0.140\linewidth,clip] {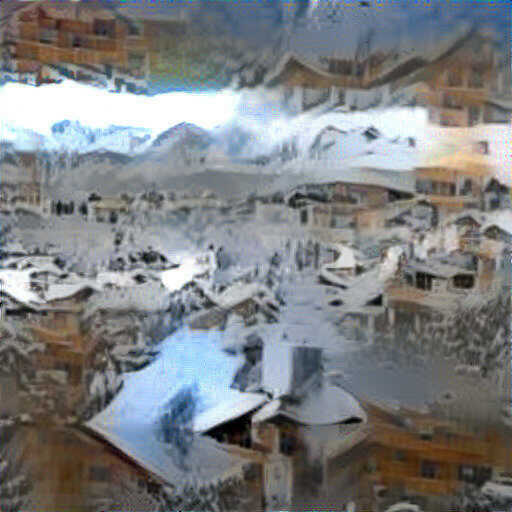}\\ 

&
\includegraphics[width=0.140\linewidth, clip]
{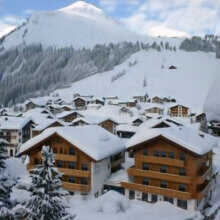}&
\includegraphics[width=0.140\linewidth,clip] {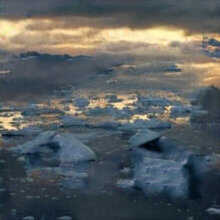}&
\includegraphics[width=0.140\linewidth,clip] 
{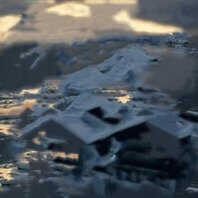}&
\includegraphics[width=0.140\linewidth,clip] {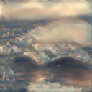}&
\includegraphics[width=0.140\linewidth,clip] {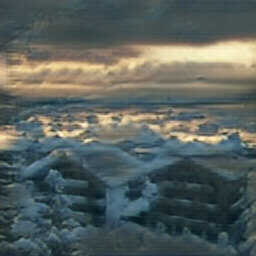}& 
\includegraphics[width=0.140\linewidth,clip] {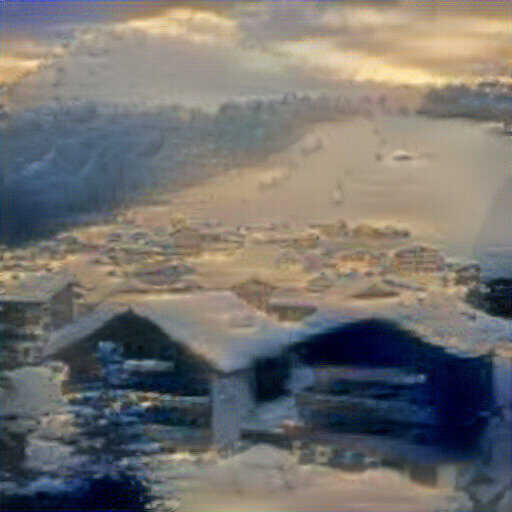}\\ 

\centering{\begin{small}\begin{turn}{90}\;\;\;\;\;\; \hspace{-18mm} HotAirBaloons2Birds \end{turn}\end{small}}&

\includegraphics[width=0.140\linewidth, clip]
{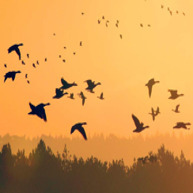}&
\includegraphics[width=0.140\linewidth,clip] {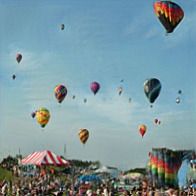}&
\includegraphics[width=0.140\linewidth,clip] 
{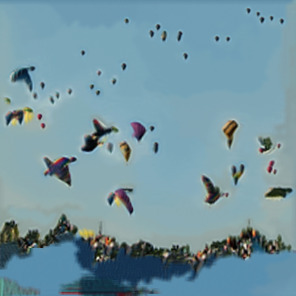}&
\includegraphics[width=0.140\linewidth,clip] {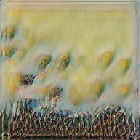}&
\includegraphics[width=0.140\linewidth,clip] {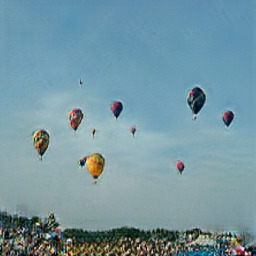}& 
\includegraphics[width=0.140\linewidth,clip] {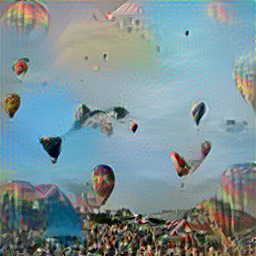}\\ 

&
\includegraphics[width=0.140\linewidth, clip]
{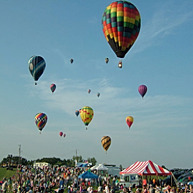}&
\includegraphics[width=0.140\linewidth,clip] {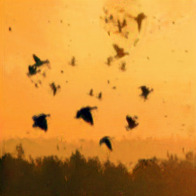}&
\includegraphics[width=0.140\linewidth,clip] 
{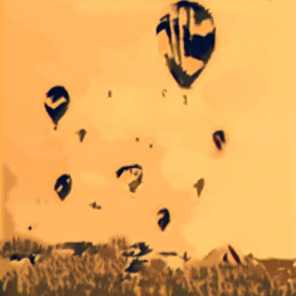}&
\includegraphics[width=0.140\linewidth,clip] {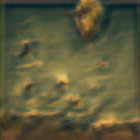}&
\includegraphics[width=0.140\linewidth,clip] {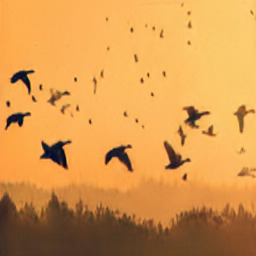}& 
\includegraphics[width=0.140\linewidth,clip] {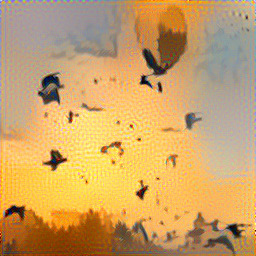}\\ 

\centering{\begin{small}\begin{turn}{90}\;\;\;\;\;\; \hspace{-18mm} Mountain2Building \end{turn}\end{small}}&

\includegraphics[width=0.140\linewidth, clip]
{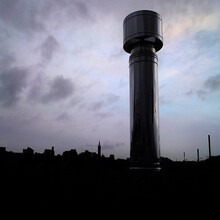}&
\includegraphics[width=0.140\linewidth,clip] {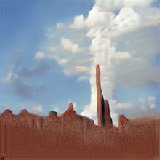}&
\includegraphics[width=0.140\linewidth,clip] 
{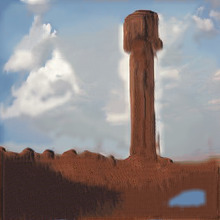}&
\includegraphics[width=0.140\linewidth,clip] {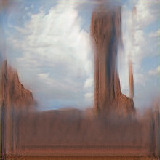}&
\includegraphics[width=0.140\linewidth,clip] {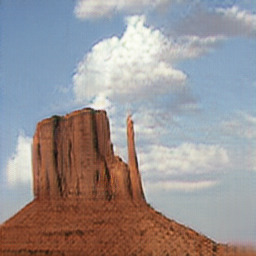}& 
\includegraphics[width=0.140\linewidth,clip] {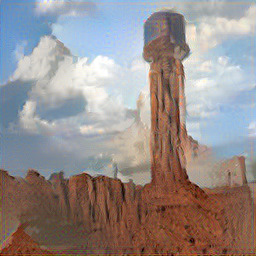}\\

&
\includegraphics[width=0.140\linewidth, clip]
{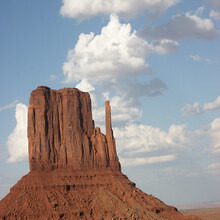}&
\includegraphics[width=0.140\linewidth,clip] {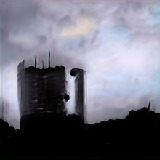}&
\includegraphics[width=0.140\linewidth,clip] 
{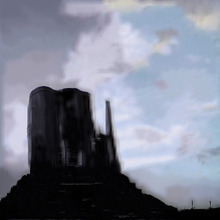}&
\includegraphics[width=0.140\linewidth,clip] {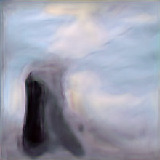}&
\includegraphics[width=0.140\linewidth,clip] {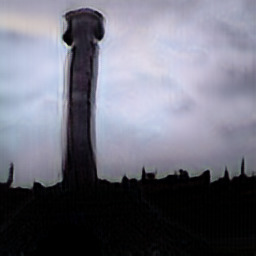}& 
\includegraphics[width=0.140\linewidth,clip] {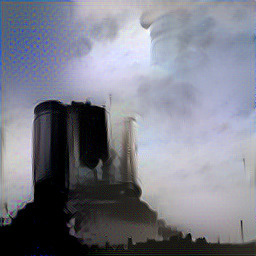}\\ 

\end{tabular}
\caption{Additional structural alignment results. 
}
\label{fig:additional3}
\end{figure*}

\begin{figure*}
  \centering
  \begin{tabular}{c@{~}c:c@{~}c@{~}c@{~}c@{~}c}
&  Input & \textbf{Ours} & DIA~\cite{deepimageanalogy} & SinGAN~\cite{singan} & Cycle$^*$~\cite{CycleGAN2017} & Style~\cite{Gatys_2016_CVPR} \\

\centering{\begin{small}\begin{turn}{90}\;\;\;\;\;\; \hspace{-18mm} Balls2Oranges \end{turn}\end{small}}&

\includegraphics[width=0.140\linewidth, clip]
{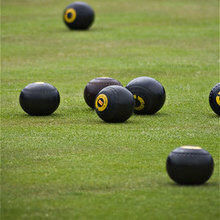}&
\includegraphics[width=0.140\linewidth,clip] {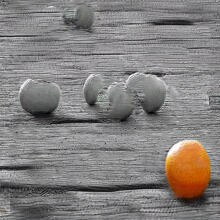}&
\includegraphics[width=0.140\linewidth,clip] 
{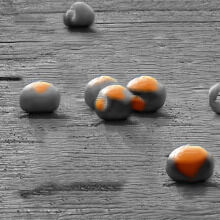}&
\includegraphics[width=0.140\linewidth,clip] {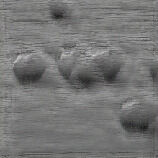}&
\includegraphics[width=0.140\linewidth,clip] {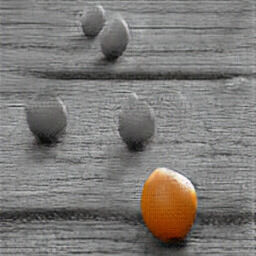}& 
\includegraphics[width=0.140\linewidth,clip] {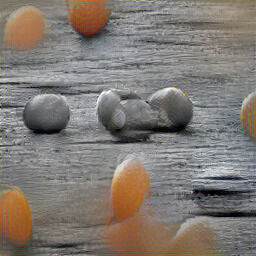}\\ 

&
\includegraphics[width=0.140\linewidth, clip]
{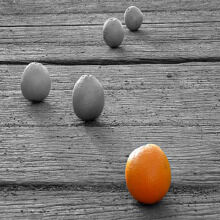}&
\includegraphics[width=0.140\linewidth,clip] {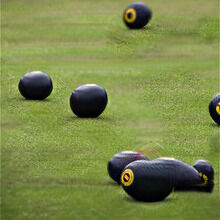}&
\includegraphics[width=0.140\linewidth,clip] 
{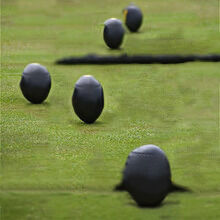}&
\includegraphics[width=0.140\linewidth,clip] {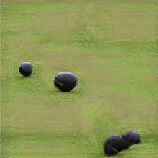}&
\includegraphics[width=0.140\linewidth,clip] {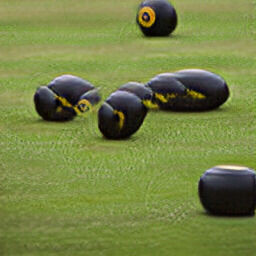}& 
\includegraphics[width=0.140\linewidth,clip] {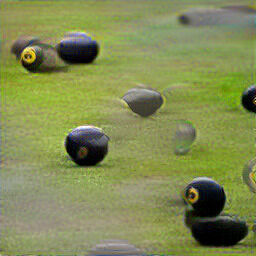}\\

\centering{\begin{small}\begin{turn}{90}\;\;\;\;\;\; \hspace{-18mm} Road2Road \end{turn}\end{small}}&

\includegraphics[width=0.140\linewidth, clip]
{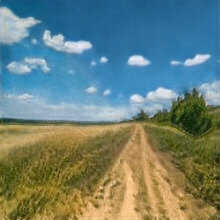}&
\includegraphics[width=0.140\linewidth,clip] {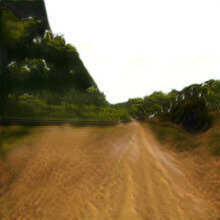}&
\includegraphics[width=0.140\linewidth,clip] 
{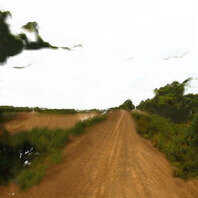}&
\includegraphics[width=0.140\linewidth,clip] {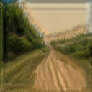}&
\includegraphics[width=0.140\linewidth,clip] {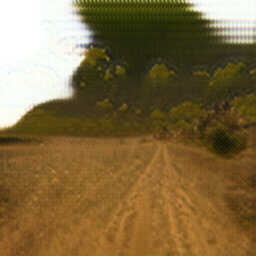}& 
\includegraphics[width=0.140\linewidth,clip] {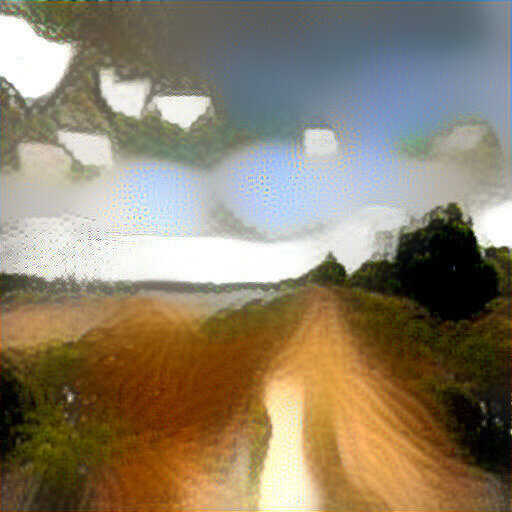}\\ 

&
\includegraphics[width=0.140\linewidth, clip]
{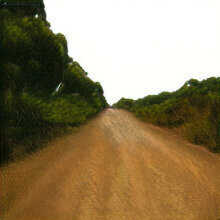}&
\includegraphics[width=0.140\linewidth,clip] {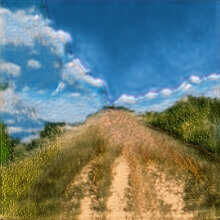}&
\includegraphics[width=0.140\linewidth,clip] 
{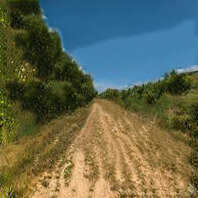}&
\includegraphics[width=0.140\linewidth,clip] {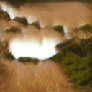}&
\includegraphics[width=0.140\linewidth,clip] {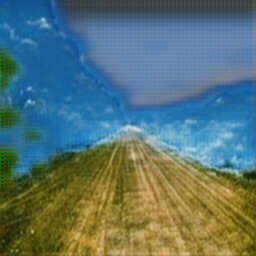}& 
\includegraphics[width=0.140\linewidth,clip] {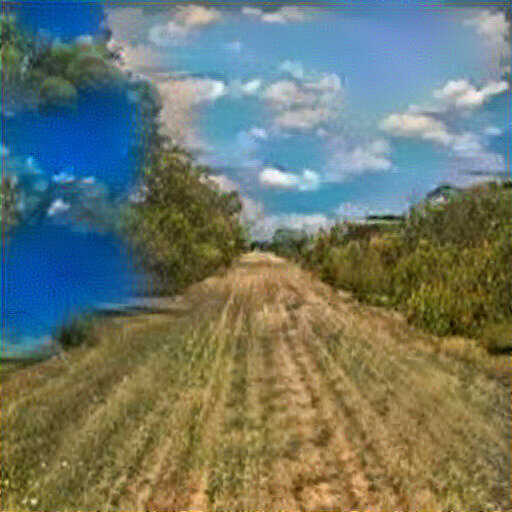}\\ 

\centering{\begin{small}\begin{turn}{90}\;\;\;\;\;\; \hspace{-18mm} Plant2Kint \end{turn}\end{small}}&

\includegraphics[width=0.140\linewidth, clip]
{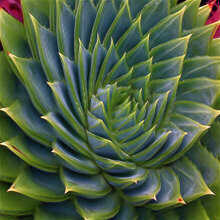}&
\includegraphics[width=0.140\linewidth,clip] {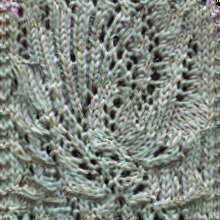}&
\includegraphics[width=0.140\linewidth,clip] 
{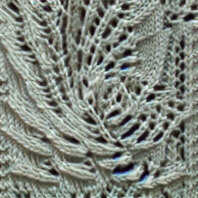}&
\includegraphics[width=0.140\linewidth,clip] {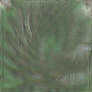}&
\includegraphics[width=0.140\linewidth,clip] {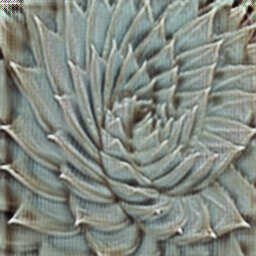}& 
\includegraphics[width=0.140\linewidth,clip] {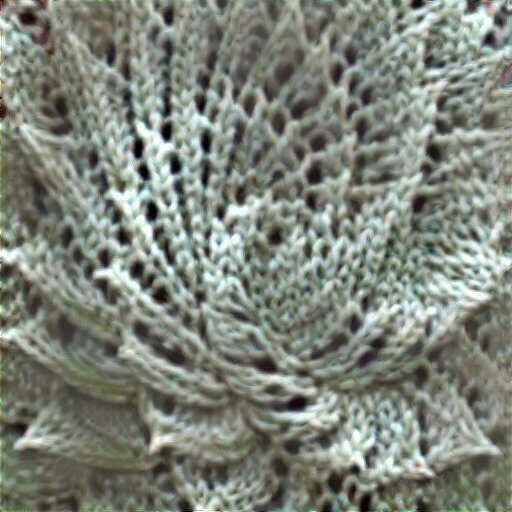}\\ 

&
\includegraphics[width=0.140\linewidth, clip]
{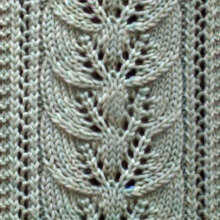}&
\includegraphics[width=0.140\linewidth,clip] {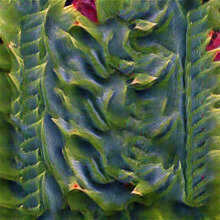}&
\includegraphics[width=0.140\linewidth,clip] 
{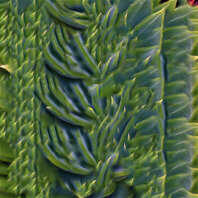}&
\includegraphics[width=0.140\linewidth,clip] {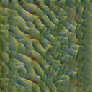}&
\includegraphics[width=0.140\linewidth,clip] {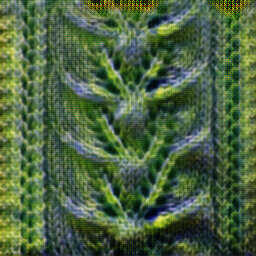}& 
\includegraphics[width=0.140\linewidth,clip] {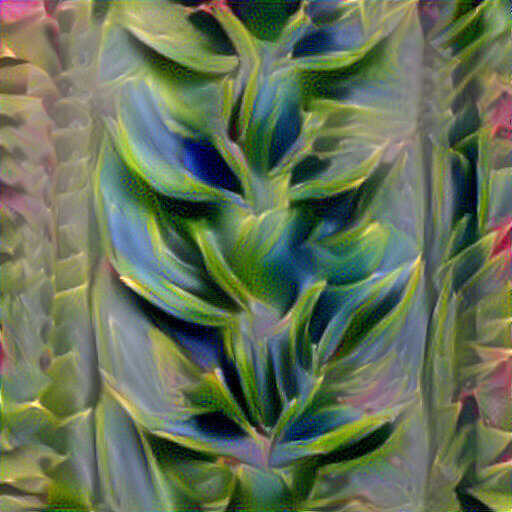}\\

\centering{\begin{small}\begin{turn}{90}\;\;\;\;\;\; \hspace{-18mm} Flowers2Flowers \end{turn}\end{small}}&

\includegraphics[width=0.140\linewidth, clip]
{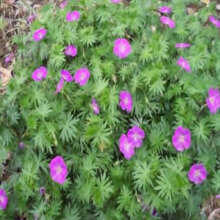}&
\includegraphics[width=0.140\linewidth,clip] {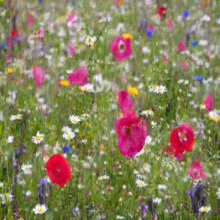}&
\includegraphics[width=0.140\linewidth,clip] 
{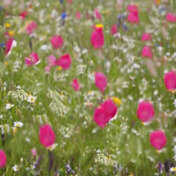}&
\includegraphics[width=0.140\linewidth,clip] {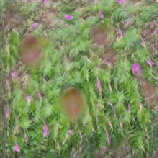}&
\includegraphics[width=0.140\linewidth,clip] {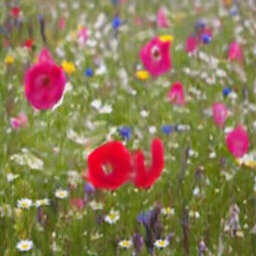}& 
\includegraphics[width=0.140\linewidth,clip] {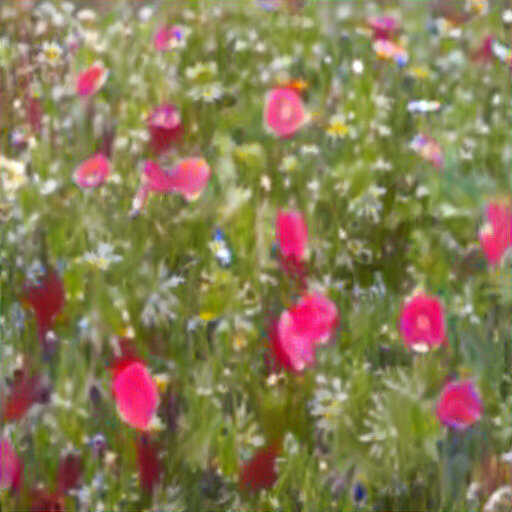}\\ 

&
\includegraphics[width=0.140\linewidth, clip]
{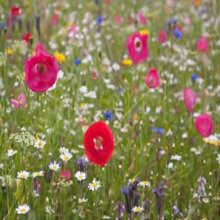}&
\includegraphics[width=0.140\linewidth,clip] {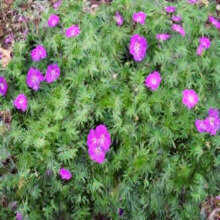}&
\includegraphics[width=0.140\linewidth,clip] 
{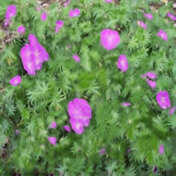}&
\includegraphics[width=0.140\linewidth,clip] {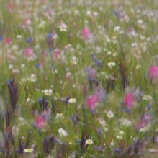}&
\includegraphics[width=0.140\linewidth,clip] {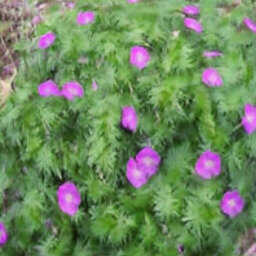}& 
\includegraphics[width=0.140\linewidth,clip] {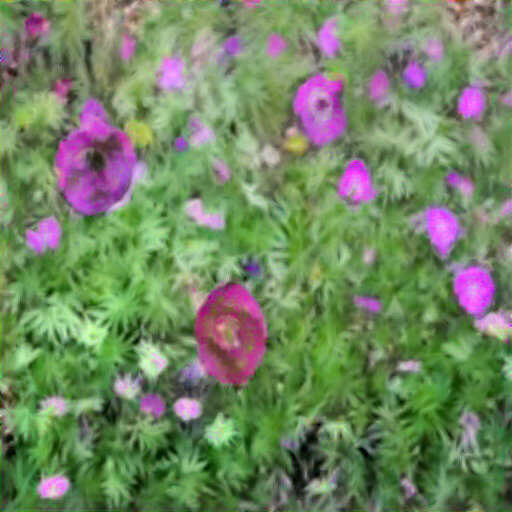}\\

\end{tabular}
\caption{Additional structural alignment results.   }
\label{fig:additional4}
\end{figure*}

\begin{figure*}
  \centering
  \begin{tabular}{c@{~}c:c@{~}c@{~}c@{~}c@{~}c}
&  Input & \textbf{Ours} & DIA~\cite{deepimageanalogy} & SinGAN~\cite{singan} & Cycle$^*$~\cite{CycleGAN2017} & Style~\cite{Gatys_2016_CVPR} \\

\centering{\begin{small}\begin{turn}{90}\;\;\;\;\;\; \hspace{-18mm} Flowers2Flowers \end{turn}\end{small}}&

\includegraphics[width=0.140\linewidth, clip]
{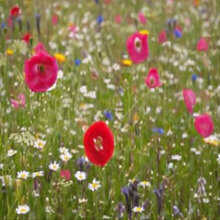}&
\includegraphics[width=0.140\linewidth,clip] {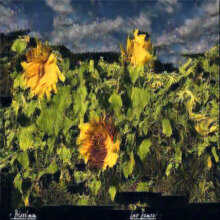}&
\includegraphics[width=0.140\linewidth,clip] 
{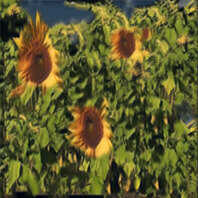}&
\includegraphics[width=0.140\linewidth,clip] {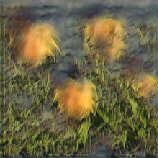}&
\includegraphics[width=0.140\linewidth,clip] {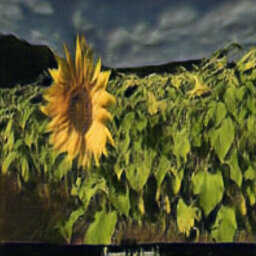}& 
\includegraphics[width=0.140\linewidth,clip] {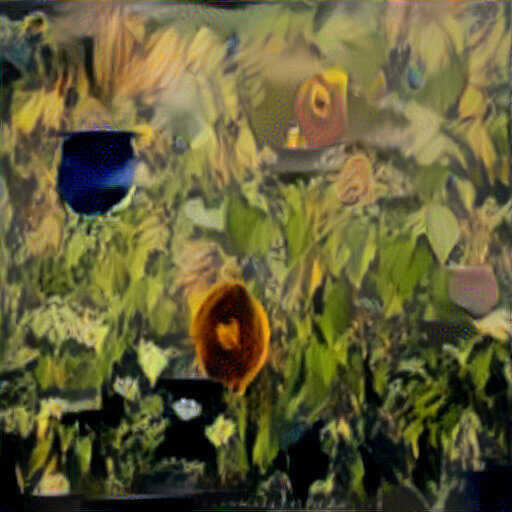}\\ 

&
\includegraphics[width=0.140\linewidth, clip]
{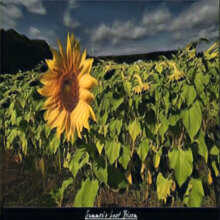}&
\includegraphics[width=0.140\linewidth,clip] {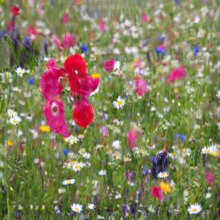}&
\includegraphics[width=0.140\linewidth,clip] 
{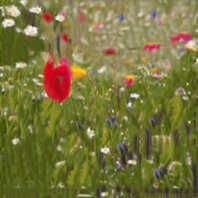}&
\includegraphics[width=0.140\linewidth,clip] {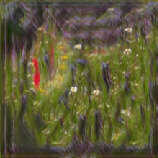}&
\includegraphics[width=0.140\linewidth,clip] {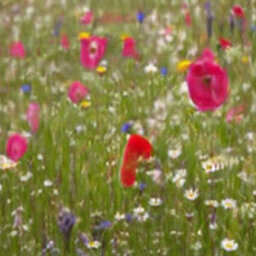}& 
\includegraphics[width=0.140\linewidth,clip] {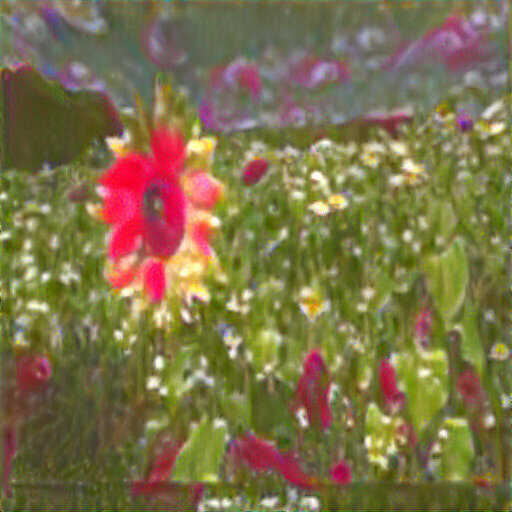}\\

\centering{\begin{small}\begin{turn}{90}\;\;\;\;\;\; \hspace{-18mm} Tennis2Marbles \end{turn}\end{small}}&

\includegraphics[width=0.140\linewidth, clip]
{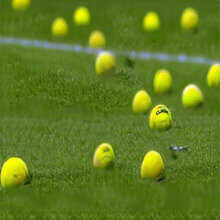}&
\includegraphics[width=0.140\linewidth,clip] {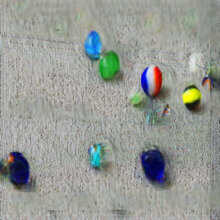}&
\includegraphics[width=0.140\linewidth,clip] 
{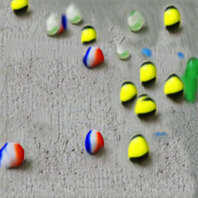}&
\includegraphics[width=0.140\linewidth,clip] {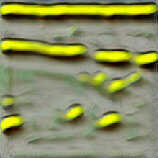}&
\includegraphics[width=0.140\linewidth,clip] {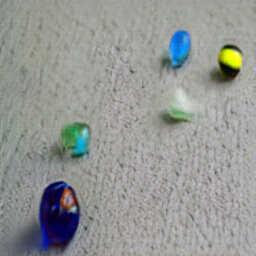}& 
\includegraphics[width=0.140\linewidth,clip] {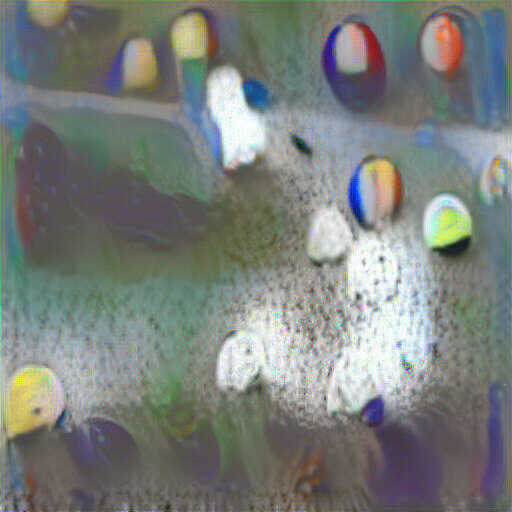}\\ 

&
\includegraphics[width=0.140\linewidth, clip]
{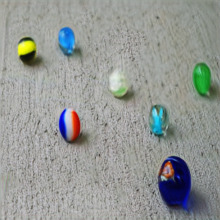}&
\includegraphics[width=0.140\linewidth,clip] {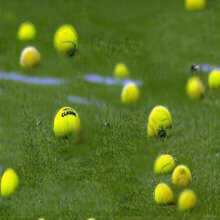}&
\includegraphics[width=0.140\linewidth,clip] 
{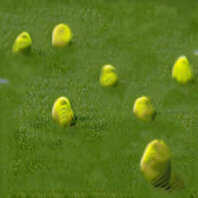}&
\includegraphics[width=0.140\linewidth,clip] {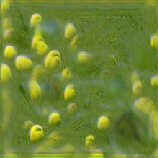}&
\includegraphics[width=0.140\linewidth,clip] {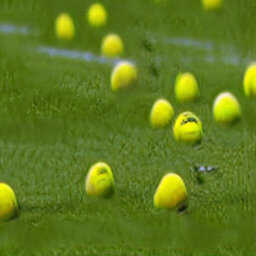}& 
\includegraphics[width=0.140\linewidth,clip] {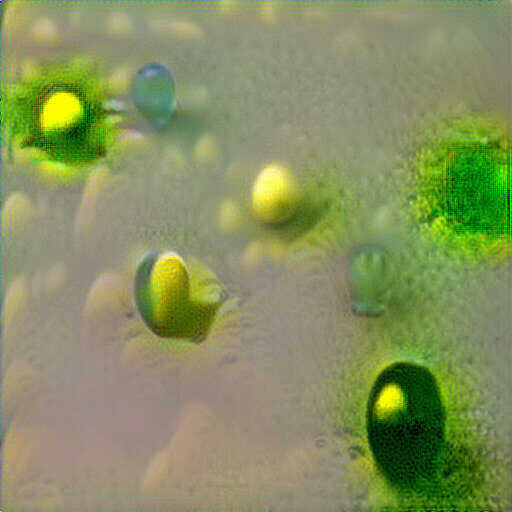}\\

\end{tabular}
\caption{Additional structural alignment results.   }
\label{fig:additional5}
\end{figure*}

\begin{figure*}
  \centering
\begin{subfigure}{0.5\textwidth}
  \centering
  \includegraphics[width=0.95\linewidth]{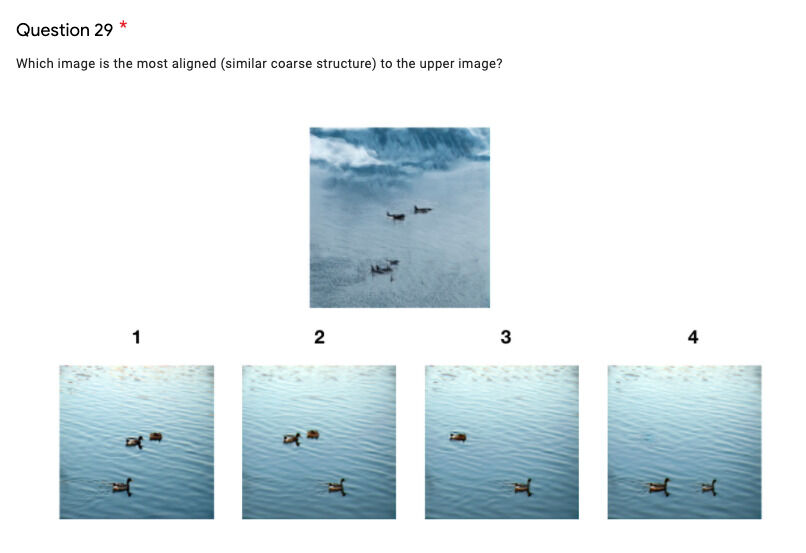}
\end{subfigure}%
\begin{subfigure}{0.5\textwidth}
  \centering
  \includegraphics[width=0.9\linewidth]{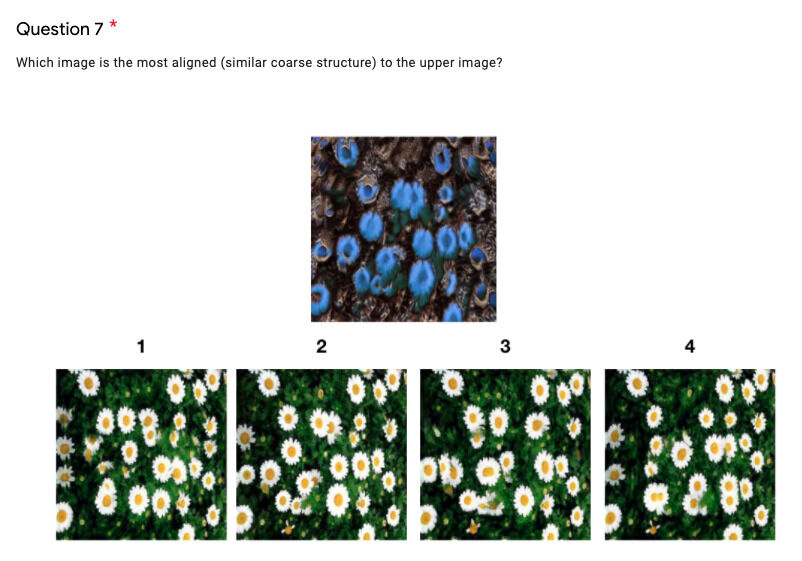}
\end{subfigure}
\caption{Screenshots from the structural alignment user study. Given the translated image $ab$ at the top, the user is requested to select the correct source image $A$ from the bottom row. }
\label{fig:c2}
\vspace{-0.1cm}
\end{figure*}

\begin{figure*}
  \centering
  \begin{tabular}{c@{~}c@{~}c:c@{~}c:c@{~}c}
 & Input & \textbf{Ours} & Input & \textbf{Ours} & Input & \textbf{Ours}\\

\centering{\begin{small}\begin{turn}{90}\;\;\;\;\;\; \hspace{-16mm} Badlands2Snow \end{turn}\end{small}}&
\includegraphics[width=0.150\linewidth, clip]
{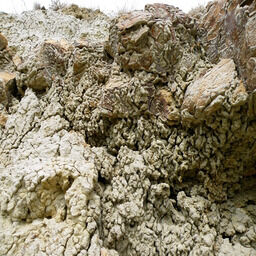}&
\includegraphics[width=0.150\linewidth,clip] {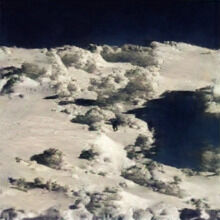}&
\includegraphics[width=0.150\linewidth,clip] 
{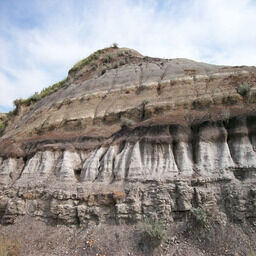}&
\includegraphics[width=0.150\linewidth,clip] {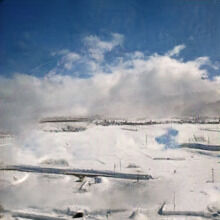}&
\includegraphics[width=0.150\linewidth,clip] 
{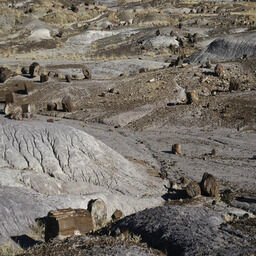}&
\includegraphics[width=0.150\linewidth,clip] {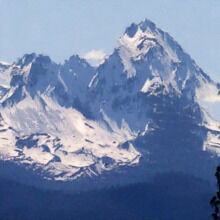}
\\ 

&
\includegraphics[width=0.150\linewidth, clip]
{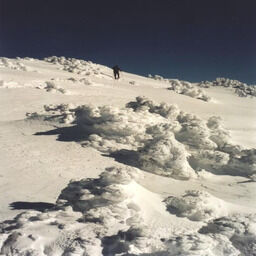}&
\includegraphics[width=0.150\linewidth,clip] {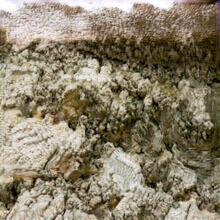}&
\includegraphics[width=0.150\linewidth,clip] 
{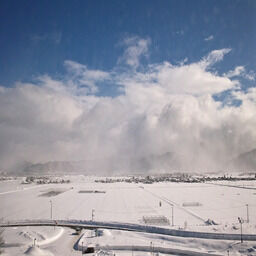}&
\includegraphics[width=0.150\linewidth,clip] {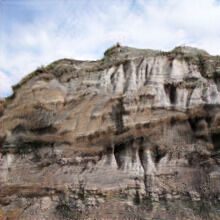}&
\includegraphics[width=0.150\linewidth,clip] 
{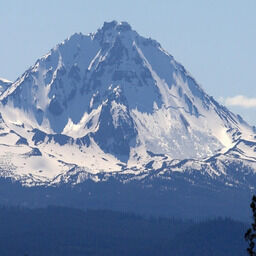}&
\includegraphics[width=0.150\linewidth,clip] {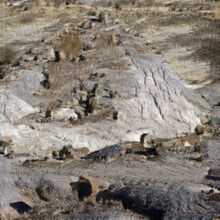}
\\ \\

\centering{\begin{small}\begin{turn}{90}\;\;\;\;\;\; \hspace{-20mm} Mountain2SnowyMountain \end{turn}\end{small}}&
\includegraphics[width=0.150\linewidth, clip]
{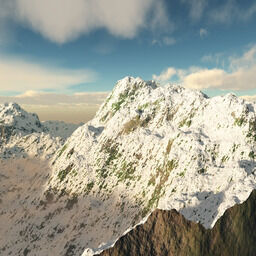}&
\includegraphics[width=0.150\linewidth,clip] {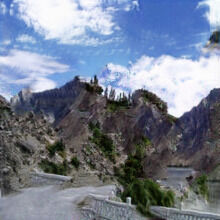}&
\includegraphics[width=0.150\linewidth,clip] 
{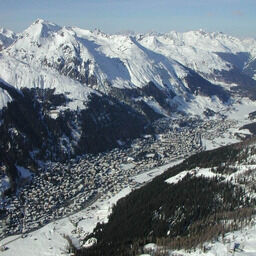}&
\includegraphics[width=0.150\linewidth,clip] {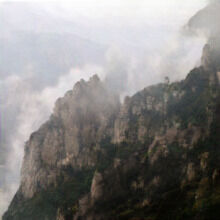}&
\includegraphics[width=0.150\linewidth,clip] 
{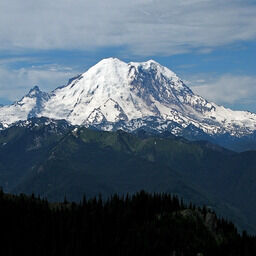}&
\includegraphics[width=0.150\linewidth,clip] {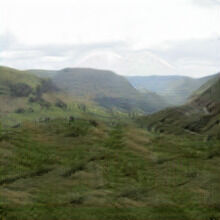}
\\ 

&
\includegraphics[width=0.150\linewidth, clip]
{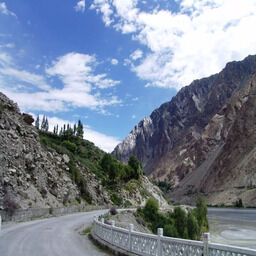}&
\includegraphics[width=0.150\linewidth,clip] {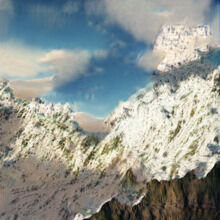}&
\includegraphics[width=0.150\linewidth,clip] 
{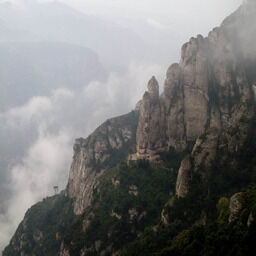}&
\includegraphics[width=0.150\linewidth,clip] {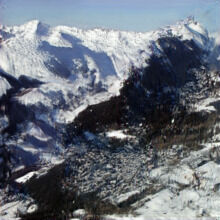}&
\includegraphics[width=0.150\linewidth,clip] 
{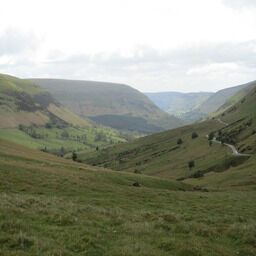}&
\includegraphics[width=0.150\linewidth,clip] {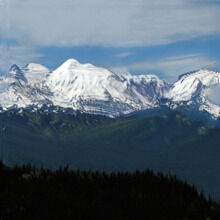}
\\ 

\end{tabular}
\caption{Random examples from the Places~\cite{places} dataset. We first chose the domains/classes (upper: "Badlands" and "Snow", lower: "mountain" and "snowy mountain") and then pick the images randomly within the domains.
}
\label{fig:rand}
\end{figure*}

\begin{figure*}
  \centering
  \begin{tabular}{c:c@{~}c@{~}c@{~}c@{~}c@{~}c@{~}c}
 Input A & $\text{K}=0$ & $\text{K}=1$ & $\text{K}=2$ & $\text{K}=3$ & $\text{K}=4$ & $\text{K}=5$ & $\text{K}=6$ \\

 \includegraphics[width=0.12\linewidth, clip] {supp/bowl_real_b.jpg}&
 \includegraphics[width=0.12\linewidth,clip] {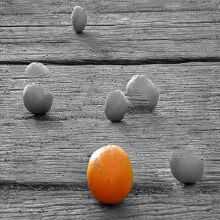}&
 \includegraphics[width=0.12\linewidth, clip] {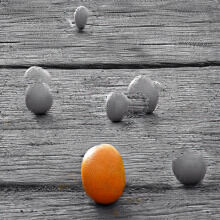}&
 \includegraphics[width=0.12\linewidth,clip] {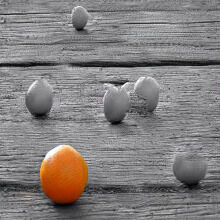}&
 \includegraphics[width=0.12\linewidth,clip] {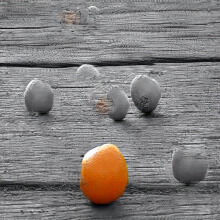}&
 \includegraphics[width=0.12\linewidth, clip] {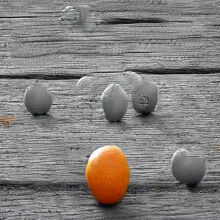}&
 \includegraphics[width=0.12\linewidth,clip] {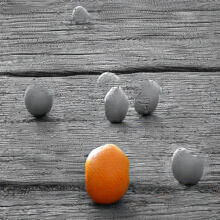}&
 \includegraphics[width=0.12\linewidth,clip] {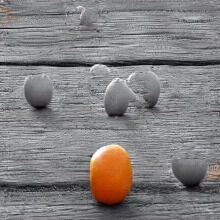}\\

 Input B & $\text{K}=7$ & $\text{K}=8$ & $\text{K}=9$ & $\text{K}=10$ & $\text{K}=11$ & $\text{K}=12$ & $\text{K}=13$ \\

\includegraphics[width=0.12\linewidth, clip] {supp/bowl_real_a.jpg}&
\includegraphics[width=0.12\linewidth,clip] {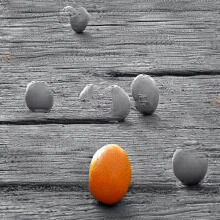}&
\includegraphics[width=0.12\linewidth, clip] {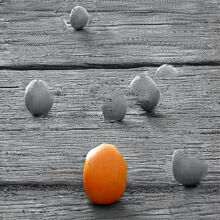}&
\includegraphics[width=0.12\linewidth,clip] {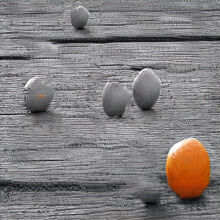}&
\includegraphics[width=0.12\linewidth,clip] {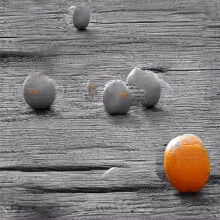}&
\includegraphics[width=0.12\linewidth, clip] {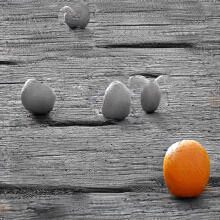}&
\includegraphics[width=0.12\linewidth,clip] {cgf_revision/k_abl2/final_a2b_k12.jpg}&
\includegraphics[width=0.12\linewidth,clip] {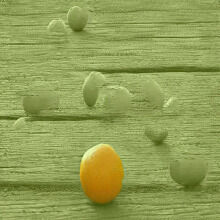}\\

\end{tabular}
\caption{ Results for a different number of residual layers $K$, out of $13$ layers, i.e. $N=12$, for the conditional generation. When $K$ is too high we get degradation in quality, while a low $K$ results in a coarser alignment. For example, for $K=0$ until $K=8$ the orange is incorrectly placed in the image. For $K=9$ until $K=12$ the orange is correctly placed.  At $K=12$ three balls appear against the three balls in the input image indicating the best alignment. At $K=13$ an incorrect texture is produced. 
}
\label{fig:k_abl2}
\end{figure*}

\begin{figure*}
  \centering
  \begin{tabular}{c:c@{~}c@{~}c@{~}c@{~}c@{~}c@{~}c}
 Input & $\text{K}=0$ & $\text{K}=1$ & $\text{K}=2$ & $\text{K}=3$ & $\text{K}=4$ & $\text{K}=5$ & $\text{K}=6$ \\

 \includegraphics[width=0.12\linewidth, clip] {supp/bowl_real_a.jpg}&
 \includegraphics[width=0.12\linewidth,clip] {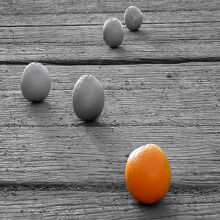}&
 \includegraphics[width=0.12\linewidth, clip] {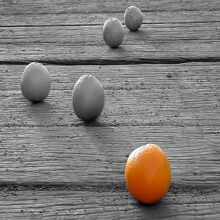}&
 \includegraphics[width=0.12\linewidth,clip] {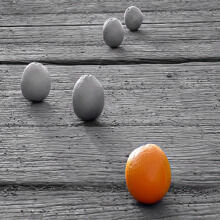}&
 \includegraphics[width=0.12\linewidth,clip] {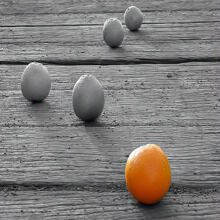}&
 \includegraphics[width=0.12\linewidth, clip] {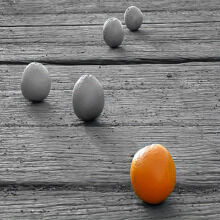}&
 \includegraphics[width=0.12\linewidth,clip] {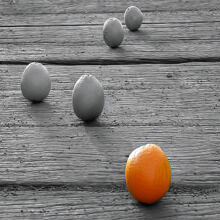}&
 \includegraphics[width=0.12\linewidth,clip] {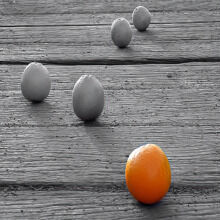}\\

  & $\text{K}=7$ & $\text{K}=8$ & $\text{K}=9$ & $\text{K}=10$ & $\text{K}=11$ & $\text{K}=12$ & $\text{K}=13$ \\

 &
\includegraphics[width=0.12\linewidth,clip] {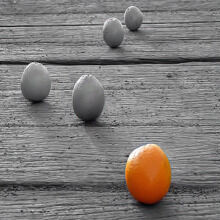}&
\includegraphics[width=0.12\linewidth, clip] {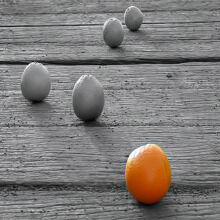}&
\includegraphics[width=0.12\linewidth,clip] {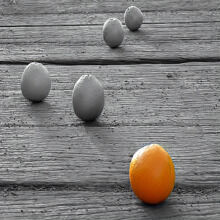}&
\includegraphics[width=0.12\linewidth,clip] {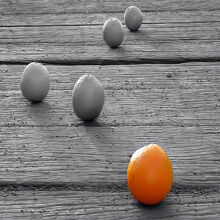}&
\includegraphics[width=0.12\linewidth, clip] {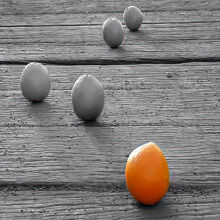}&
\includegraphics[width=0.12\linewidth,clip] {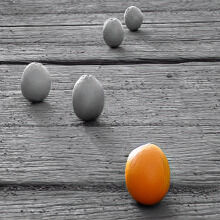}&
\includegraphics[width=0.12\linewidth,clip] {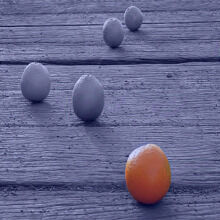}\\

\end{tabular}
\caption{Results for a different number of residual layers $K$, out of $13$ layers, i.e. $N=12$, for the unconditional generation. For each $K$, the result shown is the reconstruction of the input in the last scale $N$. As can be seen, other than $K=13$, whose texture differs from the input image, all other $K$ values result in almost perfect reconstruction. 
}
\label{fig:k_abl3}
\end{figure*}

\begin{figure*}
  \centering
  \begin{tabular}{c@{~}:c@{~}c@{~}c@{~}c@{~}c@{~}c@{~}c@{~}c@{~}c}
 Input & $S=0$ & $S=1$ & $S=2$ & $S=3$ & $S=4$ & $S=5$ & $S=6$ & $S=7$ & $S=8$ \\ 

\includegraphics[width=0.09\linewidth, clip] {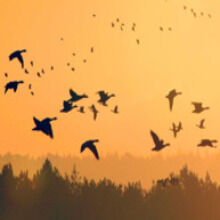}&
\includegraphics[width=0.09\linewidth,clip] {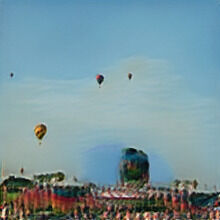}&
\includegraphics[width=0.09\linewidth, clip] {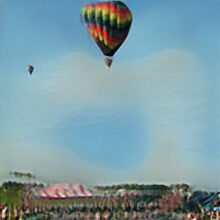}&
\includegraphics[width=0.09\linewidth,clip] {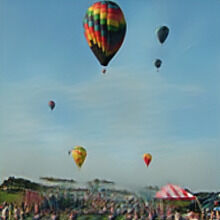}&
\includegraphics[width=0.09\linewidth,clip] {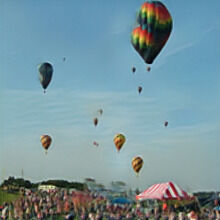}&
\includegraphics[width=0.09\linewidth, clip] {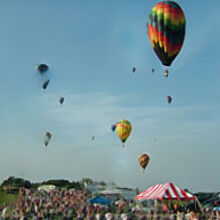}&
\includegraphics[width=0.09\linewidth,clip] {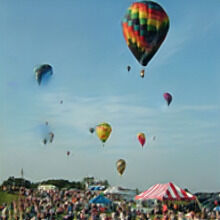}&
\includegraphics[width=0.09\linewidth,clip] {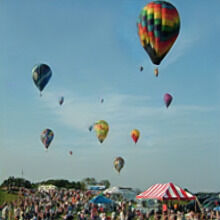}&
\includegraphics[width=0.09\linewidth,clip] {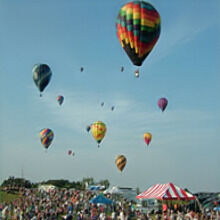}&
\includegraphics[width=0.09\linewidth,clip] {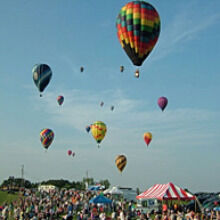} \\

\includegraphics[width=0.09\linewidth, clip] {cgf_revision/real_a_.jpg}&
\includegraphics[width=0.09\linewidth,clip] {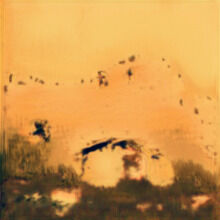}&
\includegraphics[width=0.09\linewidth, clip] {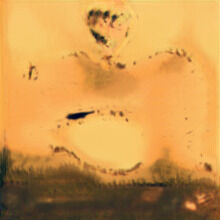}&
\includegraphics[width=0.09\linewidth,clip] {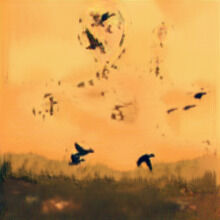}&
\includegraphics[width=0.09\linewidth,clip] {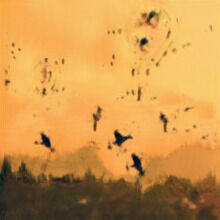}&
\includegraphics[width=0.09\linewidth, clip] {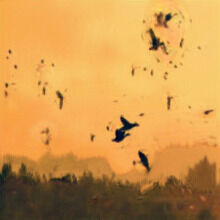}&
\includegraphics[width=0.09\linewidth,clip] {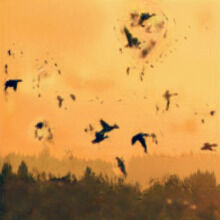}&
\includegraphics[width=0.09\linewidth,clip] {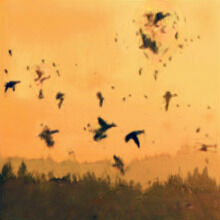}&
\includegraphics[width=0.09\linewidth,clip] {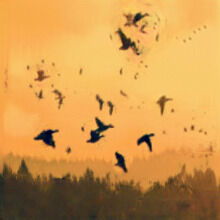}&
\includegraphics[width=0.09\linewidth,clip] {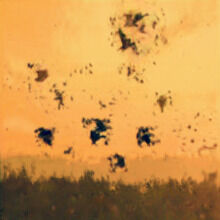} \\

\end{tabular}
\caption{The effect of injection at different scales $S$ for inference ($N=8$). The leftmost column is the input pair. For the other columns, the top row is the output $a^*_N$, when injecting $A$ at different scales $S$. The bottom row is the mapping from $a^*_N$  to $\overline{ab}_N$ at the final scale.
As can be seen, the best result is using $S=7$. For other $S$ values the image is blurrier (e.g. $S<4$), contains artifacts (e.g. $S=4,5$) or less aligned (e.g. $S=6$), i.e. different structure is present.
}
\label{fig:inj}
\end{figure*}

\begin{figure*}
  \centering
  \begin{tabular}{c@{~}c:c@{~}c@{~}c@{~}c@{~}c@{~}c}
 $A$ & $B$ & $\text{S'}=5$ & $\text{S'}=6$ & $\text{S'}=7$ & $\text{S'}=8$ & $\text{S'}=9$ & $\text{S'}=10$ \\

\includegraphics[width=0.12\linewidth, clip] {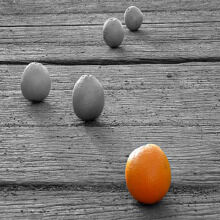}&
\includegraphics[width=0.12\linewidth,clip] {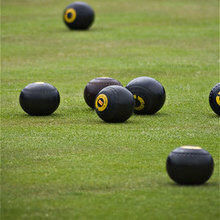}&
\includegraphics[width=0.12\linewidth, clip] {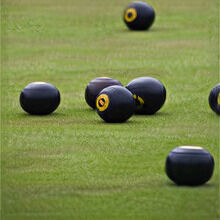}&
\includegraphics[width=0.12\linewidth,clip] {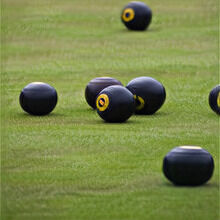}&
\includegraphics[width=0.12\linewidth,clip] {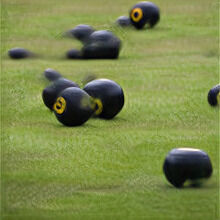}&
\includegraphics[width=0.12\linewidth, clip] {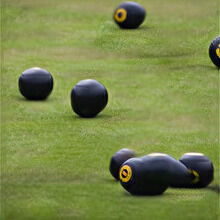}&
\includegraphics[width=0.12\linewidth,clip] {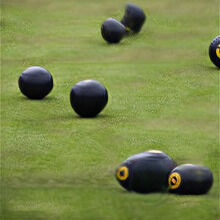}&
\includegraphics[width=0.12\linewidth,clip] {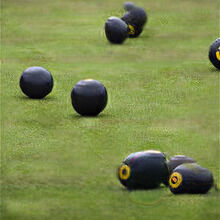}\\

\end{tabular}
\caption{Mapping from $A$ to $B$ at different scales at inference. We first inject $A$ at scale $S=5$. We then upscale $a_S$ to scale $S'$ resulting in $a^*_{S'}$. We then map $a_S'$ to the other domain resulting in $\overline{ab}_{S'}$ and then upscale $\overline{ab}_{S'}$ to $ab_N$.
As can be seen, we get better results by performing the mapping at later scales.  
}
\label{fig:switch}
\end{figure*}

\begin{figure}
  \centering
  \centering
  \begin{tabular}{c@{~}c:c@{~}c@{~}c@{~}c@{~}c}
  & Input & (100) & (1000) & (5000) & (10000)  \\
  
\centering{\begin{small}\begin{turn}{90}\;\;\;\;\;\; \end{turn}\end{small}}&
\includegraphics[width=0.17\linewidth, clip]
{image2image_results/pumpkin_real_b.jpg}&
\includegraphics[width=0.17\linewidth,clip] 
{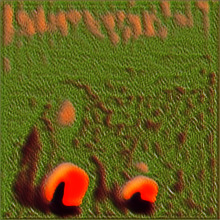}&
\includegraphics[width=0.17\linewidth,clip] 
{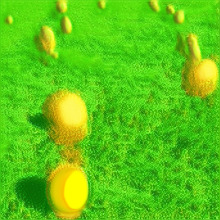}&
\includegraphics[width=0.17\linewidth,clip] 
{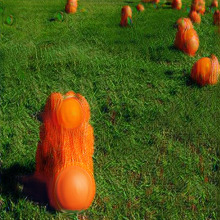}&
\includegraphics[width=0.17\linewidth,clip] 
{image2image_results/pumpkin_b2a_ours.jpg} \\

\centering{\begin{small}\begin{turn}{90}\;\;\;\;\;\; \end{turn}\end{small}}&
\includegraphics[width=0.17\linewidth,clip] {image2image_results/pumpkin_real_a.jpg}&
\includegraphics[width=0.17\linewidth,clip] 
{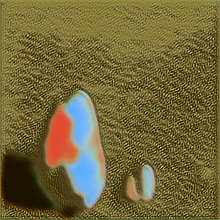}&
\includegraphics[width=0.17\linewidth,clip] 
{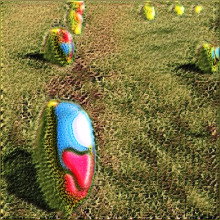}&
\includegraphics[width=0.17\linewidth,clip] 
{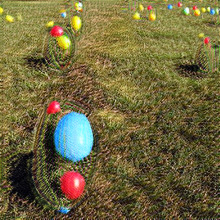}&
\includegraphics[width=0.17\linewidth,clip] 
{image2image_results/pumpkin_a2b_ours.jpg} \\

\end{tabular}
\caption{Effect of training with a different number of iterations for the translation of balls to pumpkins and vice versa. The number of iterations used for each scale is shown above the resulting output. }
\label{fig:computational_time}
\end{figure}

\begin{figure}
  \centering
  \begin{tabular}{c@{~}c:c@{~}c@{~}c@{~}c@{~}c}
 Input & \textbf{Ours} & Input & \textbf{Ours} \\

\includegraphics[width=0.2\linewidth, clip]
{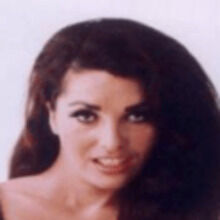}&
\includegraphics[width=0.2\linewidth,clip] {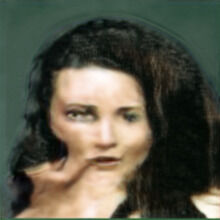}&
\includegraphics[width=0.2\linewidth, clip]
{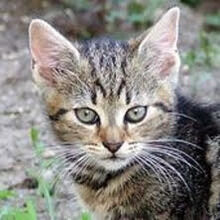}&
\includegraphics[width=0.2\linewidth,clip] {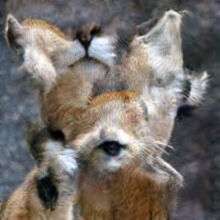}&
\\

\includegraphics[width=0.2\linewidth, clip]
{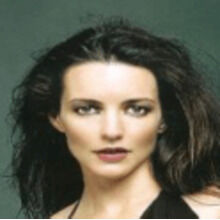}&
\includegraphics[width=0.2\linewidth,clip] {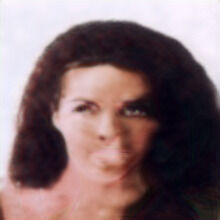}&
\includegraphics[width=0.2\linewidth, clip]
{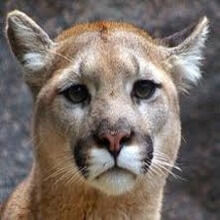}&
\includegraphics[width=0.2\linewidth,clip] {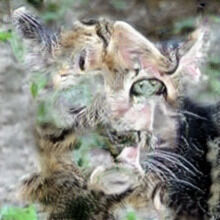}&
\\

\end{tabular}
\caption{Example failure cases of our method. See additional discussion in the main text.
}
\label{fig:fail}
\end{figure}

\end{document}